\PassOptionsToPackage{hyphens}{url}
\documentclass{article}
\usepackage{paper,times}
\usepackage[T1]{fontenc}
\usepackage{amsmath,amssymb}
\usepackage{graphicx}
\usepackage{booktabs,tabularx}
\usepackage{longtable}
\input{appendices/recipe_format}
\usepackage{microtype}
\usepackage{placeins}
\usepackage{hyperref}
\usepackage{url}
\usepackage{enumitem}
\usepackage{frontpage}
\newcommand{\method}{OpenTSLM TeeMoE}

\title{OpenTSLM \textcolor{coverAccent}{Tee}\textcolor{coverGreen}{Mo}\textcolor{coverRed}{E}: A Unified Time-Series Language Model for Forecasting, Contextual Prediction, and Reasoning}
\date{October 7, 2026}
\author{
  \mbox{Tony Chen\textsuperscript{1,2}},
  \mbox{Timo Stoffregen\textsuperscript{3}},
  \mbox{Maxwell Xu\textsuperscript{4}},
  \mbox{Thomas Kaar\textsuperscript{3,5}},\\[2pt]
  \mbox{Martin Maritsch\textsuperscript{3}},
  \mbox{Geremia Pompei\textsuperscript{6}},
  \mbox{Nicolas Zumarraga\textsuperscript{5}},
  \mbox{Robert Jakob\textsuperscript{3,5}},\\[2pt]
  \mbox{Paul Schmiedmayer\textsuperscript{2,\textdagger}},
  \mbox{Patrick Langer\textsuperscript{2,3,5,\textdagger}},
  \mbox{Juncheng Liu\textsuperscript{7,\textdagger}}
}
\coveraffiliations{
  \textsuperscript{1}Columbia University, USA\quad
  \textsuperscript{2}Stanford University, USA\quad
  \textsuperscript{3}Aionic Labs, Switzerland\\
  \textsuperscript{4}Google\quad
  \textsuperscript{5}Agentic Systems Lab, ETH Z\"urich, Switzerland\\
  \textsuperscript{6}University of Pisa, Italy\quad
  \textsuperscript{7}National University of Singapore
}

\begin{document}
\maketitle
\pagestyle{plain}
\begin{abstract}
Real-world time-series applications increasingly require models that can handle time series forecasting, context-conditioned prediction, and language-based temporal reasoning.
Yet current time-series foundation models remain fragmented across these capabilities: numerical specialists often provide the strongest forecasts, while language-based models offer broader contextual understanding and analysis. A central challenge is to unify these heterogeneous capabilities without reducing their individual performance.
We introduce \method{}, a generalist time-series language model that can forecast directly from observed time series, reason over textual context and temporal patterns, and synthesize and refine predictions from external numerical forecasting specialists. 
We independently train three low-rank experts for forecast aggregation, native forecasting, and temporal analysis over a shared backbone. A learned LoRA mixture-of-experts controller then weights their frozen parameter updates for each request.
Our proposed model achieves strong performance on widely used benchmarks for time series forecasting, context-conditioned prediction, and language-based temporal reasoning, ranking among the \textbf{top three} on GIFT-Eval by mean MASE rank, Context is Key by RCRPS, and TimeSeriesExam by accuracy.
\end{abstract}

\section{Introduction}

Time-series applications must often accommodate different types of requests.
A demand-planning system, for example, may be asked to forecast future demand, predict the effect of a planned intervention, or interpret an unusual historical pattern. This motivates a general-purpose time-series model that can handle \textbf{numerical forecasting}, \textbf{context-conditioned prediction}, and \textbf{temporal reasoning}, while retaining the accuracy that makes specialized systems useful.

These capabilities, however, have traditionally been addressed by different model families.
Numerical time-series foundation models (TSFMs) learn to forecast across datasets through large-scale pretraining~\citep{timesfm,chronos,moirai}, while time-series language models (TSLMs) connect observations with instructions, context, and analytical questions~\citep{opentslm,chatts}.
Context can describe an intervention or constraint that is absent from the observed history; reasoning may instead require comparing signals or identifying a temporal relationship.
Recent general-purpose temporal models bring multiple numerical or language-based tasks into a shared model~\citep{units,tsllm,timeomni,timeomnivl}.
Their capabilities range from forecasting, classification, and imputation to answering questions about signals and generating time series from language.
However, broader task coverage does not necessarily preserve the performance of strong task-specific models.
This raises a central question: \textit{can a general-purpose model acquire broad capabilities without diluting specialist performance}? 

One way to address this challenge is through a mixture of experts (MoE): learn the capabilities separately, then integrate them within a shared backbone.
LoRA-library composition has been shown to largely retain specialist performance across training tasks~\citep{ostapenko2024modular}. This modular design lets each specialist use its own objectives, optimization settings, and batching strategies, and be revised without retraining the others.

We follow this approach in \method{} (\textbf{T}im\textbf{e}s\textbf{e}ries \textbf{M}ixture \textbf{o}f \textbf{E}xperts), a generalist time-series language model that combines numerical forecasting, context-conditioned prediction, and temporal reasoning over a shared Qwen3.6-27B backbone~\citep{qwen36_27b}.
We independently train three low-rank capability adapters~\citep{lora}, each with its own data and objective.
For numerical forecasting, a \emph{foundation model aggregation expert} refines predictions assembled from external forecasters.
For context-conditioned prediction, a \emph{native forecasting expert} learns to generate future values directly from historical observations and textual context, and for temporal reasoning, a \emph{temporal analysis expert} learns to answer analytical questions about observed signals.

\begin{figure*}[htbp]
\centering
\includegraphics[width=\linewidth]{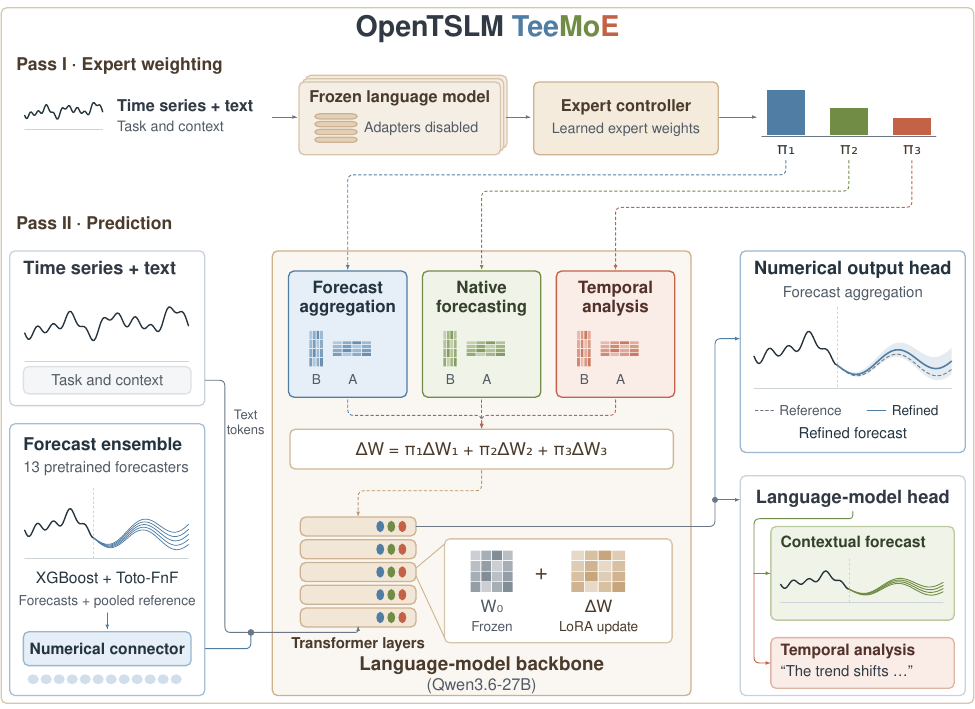}
\caption{\method{} shares one language backbone across three capability adapters. An adapter-disabled pass forms the request representation for the controller, which selects adapter weights and an output path. A second pass applies the mixed adapters to the full expert inputs. When the aggregation expert has a mixture weight above 0.5, external forecasts enter through the numerical connector. Dashed arrows denote mixture weights.}
\label{fig:system}
\end{figure*}

Following this capability-preserving approach, we freeze the experts and train a controller to assign their weights and select the output path for each request (Figure~\ref{fig:system}).

Our contributions are:
\begin{itemize}[leftmargin=*,beginpenalty=10000]
  \item We introduce a generalist time-series language model for numerical forecasting, context-conditioned prediction, and temporal reasoning, built from independently trained experts within a shared backbone. To our knowledge, this is the first use of MoE to construct a generalist TSLM. Code and model checkpoints are publicly available.
  \item We show that a single generalist model can achieve competitive performance across benchmarks for different capabilities where specialized models usually dominate. As of September 25, 2026, \method{} ranks among the top three on GIFT-Eval, Context is Key, and TimeSeriesExam~\citep{gifteval,contextiskey,timeseriesexam}.
  \item We use TeeMoE to study how mixture-of-experts composition affects TSLM capabilities and training. Comparisons of specialists, composition, and joint training characterize capability preservation, expert interaction, and training cost, guiding future generalist TSLMs.
\end{itemize}

\section{Related Work}
\label{sec:related}

\paragraph{Time-series foundation models.}
Large-scale pretraining has shifted numerical forecasting toward reusable models that transfer across datasets and domains.
TimesFM, Chronos, Moirai, Timer, Toto, and Lag-Llama develop this approach with different architectures and probabilistic representations~\citep{timesfm,chronos,moirai,timer,toto,lagllama}.
By reusing temporal patterns learned across many series, such models can infer the underlying dynamics from the input context and extend them to provide strong forecasts on new datasets with little or no task-specific training~\citep{timesfm,chronos}.

A model with the best average forecast error need not be the most accurate on every series or at every point in the horizon~\citep{synapse}.
Forecast combination exploits this variation: FFORMA learns ensemble weights from time-series features~\citep{fforma}, and Chroma selects or ensembles a portfolio of pretrained specialists using validation forecasts~\citep{chroma}.
Among ensembles of existing foundation models, Synapse adapts mixture weights across the horizon using a rolling performance estimate and simulated observations~\citep{synapse}; CastStar's released system uses a learned gate to select or weight candidate forecasts~\citep{caststar}.
Our aggregation expert extends this forecast-level interface by exposing both an ensemble prediction and its constituent distributions to a language backbone for refinement.

\paragraph{Language models for prediction and understanding.}
Adapting language models to time series offers access to pretrained representations as well as a natural-language interface.
GPT4TS and Time-LLM exploit pretrained transformer representations for numerical tasks~\citep{gpt4ts,timellm}.
PromptCast formulates forecasting as sentence-to-sentence prediction~\citep{promptcast}, and LLMTime expresses it as continuation of numerical text~\citep{llmtime}.
For temporal understanding, OpenTSLM and ChatTS connect encoded observations to language-based interpretation~\citep{opentslm,chatts}.
TS-Reasoner aligns a frozen time-series foundation model with an LLM through time-series--caption pretraining followed by instruction tuning~\citep{tsreasoner}.
Coding agents instead generate and execute Python analyses to answer questions about time series~\citep{codingagents}.

Text becomes part of the forecasting problem when it describes an event, constraint, or relationship that cannot be inferred from history alone.
Context is Key makes this requirement explicit through forecasting tasks with essential contextual information~\citep{contextiskey}.
Beyond Na\"ive Prompting studies both direct forecast generation and contextual modification of numerical forecasts~\citep{naiveprompting}; LLM as Forecasting Planner uses language-model planning to guide a numerical forecaster~\citep{lafp}.

\paragraph{General-purpose temporal models.}
General-purpose temporal models handle multiple time-series tasks within a single model.
UniTS brings forecasting, classification, imputation, and anomaly detection into a shared numerical model~\citep{units}.
MOMENT learns reusable time-series representations through large-scale pretraining and adapts them to downstream tasks with limited supervision~\citep{moment}.
TsLLM extends language-based modeling across forecasting and time-series question answering~\citep{tsllm}, and TimeOmni-1 connects temporal perception with reasoning and decision making~\citep{timeomni}.
TimeOmni-VL pursues understanding and generation through a visual time-series interface~\citep{timeomnivl}.

\paragraph{Expert specialization and composition.}
LoRA provides low-rank adaptations~\citep{lora}, used by TEMPO for forecasting and OpenTSLM for temporal reasoning~\citep{tempo,opentslm}.
AdapterFusion separates task-specific learning from composition, reusing frozen adapters over a shared backbone~\citep{pfeiffer-etal-2021-adapterfusion}.
Modular LoRA libraries demonstrate the potential for capability retention: Arrow routing nearly recovers the average performance of oracle specialist selection across training tasks using a task-specific library~\citep{ostapenko2024modular}.
MoLE likewise learns layer-specific gates over frozen LoRAs to preserve their individual characteristics~\citep{wu2024mole}.
TeeMoE follows this modular-library perspective: experts learn distinct temporal capabilities, and composition makes them available within one generalist.
We study this retention alongside independent expert training and revision.
Our controller uses request-level weights shared across layers, drawing its two-pass execution from X-LoRA~\citep{xlora}.

\section{OpenTSLM TeeMoE}
\label{sec:method}
\method{} connects independently trained aggregation, native forecasting, and analysis adapters through a shared backbone and learned controller (Figure~\ref{fig:system}).

\subsection{Design and request interface}

A request contains an observed series $x_{1:T}$, a task instruction, and any context or question.
The model first chooses its expert mixture, then computes the response (Figure~\ref{fig:system}).
In the first pass, the backbone runs with all adapters disabled to form the request representation for the controller.
The controller uses this representation to assign weights to the three adapters and choose the numerical decoder or language-model head.
In the second pass, the backbone uses this adapter mixture to refine the ensemble forecast through the numerical decoder or generate a forecast or analytical answer through the language-model head.
Appendix~\ref{app:shared-prompt} gives the shared instruction and request templates.

\subsection{The foundation model aggregation expert}

Pretrained forecasting models offer a source of temporal knowledge for language models: OpenTSLM explores a Chronos-2 encoder~\citep{opentslm,chronos2}, and the ARFBench hybrid connects Toto representations to a vision-language model~\citep{arfbench,toto}.
Our aggregation expert builds a reference forecast by ensembling pretrained forecasters, then uses the shared backbone to refine it from the candidates' agreement and disagreement.

\paragraph{Constructing the reference forecast.}
An XGBoost regressor predicts the relative forecast ranks of eight selected models from the observed history and candidate predictions~\citep{xgboost}.
Its predicted rank scores are converted into weights for pooling the candidates' cumulative distribution functions (CDFs) into a quantile forecast.
To broaden model coverage, we blend it with Toto-FnF, a public ten-model ensemble also used as a forecasting component in RacineCast-1~\citep{totofnf,racinecast}, using a scalar weight learned from training data.
The two model pools share five forecasters, so their union contains thirteen candidates.
The reference quantiles are $q_{0,t,\tau}$ at future step $t$ and quantile level $\tau$.
Appendices~\ref{app:numerical-interface} and~\ref{app:aggregation-recipe} give the model roster and exact pooling and blending rules.

\paragraph{Encoding numerical evidence.}
The numerical encoder summarizes forecast levels, uncertainty, and candidate disagreement at a fixed set of horizon positions.
It normalizes differences relative to the reference and maps this evidence into continuous input tokens for the shared language backbone.
The decoder reads the resulting hidden states and interpolates its outputs over the requested horizon.

\paragraph{Refining the reference.}
The decoder refines forecast location while preserving quantile ordering and interval widths.
It predicts a correction $d_t$, giving the final forecast
\begin{equation}
 q_{t,\tau}=q_{0,t,\tau}+g_t s_t d_t,
 \qquad |d_t|\leq\tfrac12,\quad 0\leq g_t\leq1.
 \label{eq:edit}
\end{equation}
Here $s_t$ sets the correction scale in signal units, and the disagreement gate $g_t$ reduces the edit when candidate disagreement is large relative to reference uncertainty.
The common shift can change coverage by correcting location bias.
A zero-initialized decoder starts from the reference forecast, so training learns a residual adjustment.
Appendix~\ref{app:numerical-interface} specifies the encoder, decoder, and correction functions; Appendix~\ref{app:aggregation-recipe} describes the XGBoost reference when Toto-FnF is unavailable.

\subsection{The native forecasting expert}

The native expert forecasts directly from history and textual context, without external forecasters.
Its request contains timestamp--value pairs, the future timestamps, and any context describing interventions, constraints, or relationships.
The backbone's language-model head generates timestamp--value pairs covering the requested horizon.
Each response is one possible future trajectory; sampling multiple responses yields an empirical predictive distribution.
Appendix~\ref{app:shared-prompt} gives the formats.

\subsection{The temporal analysis expert}

The analysis expert identifies patterns and anomalies, compares series, and interprets statistical relationships.
Analysis requests contain observations, a question, and any supplied options or definitions, using the evidence format in Appendix~\ref{app:analysis-prompts}.
The language-model head generates the selected option's label and text for multiple-choice questions, or a free-form answer.

\subsection{Learned expert composition}
\label{sec:composition}

To retain the specialized capabilities through modular LoRA composition~\citep{ostapenko2024modular}, we freeze the backbone, all three adapters, and the aggregation components, and train only the controller.
The controller maps the request representation to three softmax weights $\pi$.
Appendix~\ref{app:output-selection} defines this representation, and Appendix~\ref{app:routing-example} illustrates the request format.

For an adapted matrix in layer $\ell$, composition applies
\begin{equation}
 W'_\ell=W_\ell+\sum_{e=1}^{3}\pi_e\Delta W_{e,\ell},
 \qquad \pi_e\geq0,\quad\sum_e\pi_e=1.
 \label{eq:xlora}
\end{equation}
Here $W_\ell$ is a frozen backbone matrix and $\Delta W_{e,\ell}$ is expert $e$'s low-rank update, including its fixed LoRA scaling.
One mixture is shared across adapted layers and held fixed for the entire response.
These weights combine parameter updates; XGBoost weights combine forecast distributions.

The learned aggregation weight selects the forecast decoder when $\pi_{\mathrm{agg}}>0.5$, and token generation otherwise; no context/no-context switch is hard-coded.
Both output paths use the mixed adapters; the numerical path obtains candidate forecasts from the observed history.

\section{Training}
\label{sec:training}

We fit the numerical ensemble, train each capability expert with its own data and objective, then train the controller over the frozen experts.
The shared pretrained backbone remains frozen throughout.
Appendix~\ref{app:architecture} provides adapter configurations and optimization settings.

\subsection{Aggregation training}

We fit the numerical ensemble on 576,920 forecasting windows spanning economics, energy, healthcare, environmental monitoring, retail, transport, and computing systems.
XGBoost predicts candidate ranks under empirical continuous ranked probability score (CRPS).
We also fit the scalar blend of XGBoost and Toto-FnF forecasts.
Appendix~\ref{app:aggregation-recipe} details the sources and fitting.

We jointly train the aggregation adapter, numerical encoder, and forecast decoder on 4,096 examples sampled from the GIFT training split.
The editor learns a bounded correction from the reference median to the observed future using a weighted SmoothL1 loss and a squared-edit penalty.
The loss supervises $d_t$ in Equation~\ref{eq:edit}; the disagreement gate is applied at inference.
Appendix~\ref{app:aggregation-recipe} specifies reference construction, sampling, and loss normalization.

\subsection{Native forecasting training}

The native expert trains for one epoch on 20,000 contextual forecasting examples drawn from real and synthetic time series.
Training uses teacher-forced cross-entropy on the response tokens.
Forward Kullback--Leibler (KL) regularization limits changes from the frozen base model's next-token distribution.
Appendix~\ref{app:native-recipe} details the data mixture, augmentation, and optimization.

\subsection{Temporal analysis training}

The analysis expert trains for one epoch on 12,000 time-series question-answering examples with categorical and free-form targets.
The analysis adapter learns through response-token cross-entropy, with forward and reverse KL regularization to the frozen base model.
Appendix~\ref{app:analysis-recipe} gives the data mixture and optimization settings; Appendix~\ref{app:analysis-prompts} describes the shared training and inference prompt format.

\subsection{Controller fitting}

The controller is a linear map from the request representation to three logits followed by a softmax.
We train it for one epoch on 1,000 examples sampled from the expert-training populations.

Training combines capability prediction losses with numerical-versus-textual output supervision.
Each example's capability determines its training head; inference uses the learned output switch.
Let $\mathcal L_c$ be the prediction loss for capability $c$, and $\mathcal H_c$ its output-format loss. Each update minimizes
\begin{equation}
 \mathcal L_{\mathrm{ctrl}}=\frac13\sum_c(\mathcal L_c+\mathcal H_c).
 \label{eq:controller-training}
\end{equation}
Textual capabilities use response-token cross-entropy.
Aggregation uses the normalized-correction loss without the squared-edit penalty.
The format loss is the example average of $-\log\pi_{\mathrm{agg}}$ for numerical examples and $-\log(1-\pi_{\mathrm{agg}})$ for text examples.
It supervises the output mechanism while the prediction losses train the adapter mixture.
Appendix~\ref{app:composition-recipe} gives the complete recipe.

\section{Experiments}
\label{sec:experiments}

We evaluate TeeMoE across numerical forecasting, context-conditioned prediction, and temporal reasoning.
Comparisons with published systems establish its benchmark standing; ablations examine individual experts, ensembling and refinement, and expert combination.

\subsection{Benchmarks and comparisons}

\paragraph{GIFT-Eval.}
We use 97 evaluation cells spanning datasets, frequencies, and horizons, comprising 371,330 forecast windows~\citep{gifteval,gifteval_leaderboard}.
Our primary metric follows the official leaderboard: mean rank by mean absolute scaled error (MASE; lower is better).

\paragraph{Context is Key (CiK).}
CiK measures probabilistic forecasting with essential textual context.
We report region-of-interest continuous ranked probability score (RCRPS; lower is better) on all 355 instances spanning 71 task types~\citep{contextiskey}.
Following the original CiK evaluation protocol~\citep{contextiskey}, we use 25 forecast trajectories per instance for our model and local comparisons.
Numerical foundation models receive the observed history without textual context.

\paragraph{TimeSeriesExam (TSE).}
TSE tests temporal understanding through 746 questions in its official v1.1 release; we report answer accuracy~\citep{timeseriesexam,tsev11}.
Our model receives the observed series, numerical summaries, and any supplied question definitions or clarifications.

We exclude reports with identified train--test leakage, incompatible evaluation populations, or an unavailable ranking metric.
Appendices~\ref{app:leaderboard-audit} and~\ref{app:baseline-evaluation} document comparison sources, ranking procedures, and model-specific evaluation settings.
For expensive comparisons, OOT denotes a projected evaluation cost above a 1,000 H100 GPU-hour budget, using the procedure in Appendix~\ref{app:gift-runtime}.

\subsection{Performance across capabilities}

\begin{table*}[t]
\centering
\caption{The same composed \method{} model across three benchmarks. DP: direct prompting; CorDP: direct prompting for forecast correction; SW: SampleWise. TSFMs receive no textual context on CiK; n/a denotes unsupported output. * marks our evaluations; OOT denotes projected cost above our compute budget (Appendix~\ref{app:gift-runtime}). TimeOmni-VL's TSE score likely reflects unsuccessful transfer despite using its official reasoning interface and the TSE question and scoring protocol (Appendix~\ref{app:timeomni-tse}). Comparison sources appear in Appendix~\ref{app:leaderboard-audit}.}
\label{tab:leaderboards}
\includegraphics[width=\linewidth]{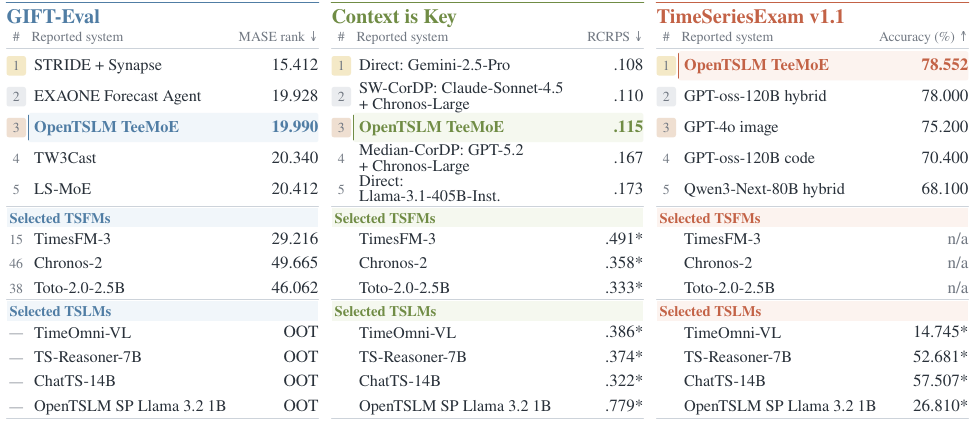}
\end{table*}

Table~\ref{tab:leaderboards} compares one composed \method{} checkpoint across all three benchmarks with leading results from the official GIFT-Eval leaderboard and published CiK and TSE evaluations.
Published systems retain their reported inputs, inference settings, and tool access.
Placements follow the ordering of benchmark scores.
TeeMoE ranks among the top three on all three benchmarks, with a mean MASE rank of 19.990 on GIFT-Eval, an RCRPS of 0.115 on CiK, and 78.552\% accuracy on TSE.

These results show that broad task coverage can coexist with strong predictive performance.
TeeMoE outperforms the selected individual numerical foundation models on GIFT, achieves lower CiK RCRPS than the locally evaluated time-series language models under the documented transfer protocols, and leads the TSE comparison.

This performance is achieved with approximately 40B total parameters, including a 27B language backbone, external forecasting models, and learned adapters, encoder, and decoder.
Despite this smaller total size, TeeMoE outperforms Llama-3.1-405B-Instruct (405B parameters) on CiK and the GPT-oss-120B hybrid agent (approximately 117B parameters) on TSE~\citep{llama3,gptoss}.
It also surpasses GPT-5.2-based forecast correction on CiK.

\subsection{Understanding TeeMoE's performance}
\label{sec:analysis}

We examine how TeeMoE's modular design supports its performance across the three capabilities.
The comparisons distinguish specialist retention from gains through expert interaction, and assess the training costs of separate specialization and joint optimization.
Tables~\ref{tab:individual-experts}--\ref{tab:numerical-baselines} compare experts, adapter combinations, and ensembling and refinement; Appendix~\ref{app:capability-ablations} gives the protocols.

\subsubsection{Individual experts}

We first compare TeeMoE with each expert activated individually to assess how combining them affects performance across capabilities.

\begin{table}[htbp]
\centering
\caption{Individual experts and TeeMoE. Deltas measure improvements over the pre-adaptation baseline: the pre-edit ensemble on GIFT and prompted unadapted Qwen on CiK and TSE. Parentheses in the baseline row give absolute scores. The aggregation expert uses its numerical interface on GIFT and its adapter with the language-model head on CiK and TSE. OOT follows Appendix~\ref{app:gift-runtime}.}
\label{tab:individual-experts}
\includegraphics[width=\linewidth]{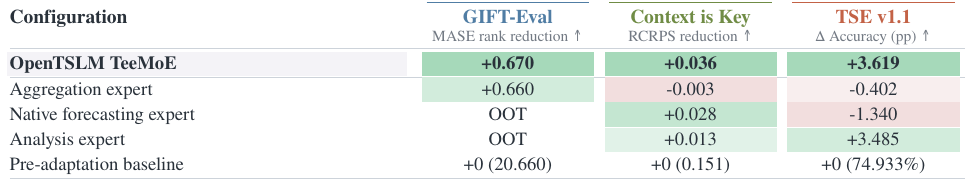}
\end{table}

Unadapted Qwen3.6-27B already provides a strong starting point, but TeeMoE reduces CiK RCRPS by 0.036 and raises TSE accuracy by 3.619 percentage points with the same inputs.
The composed system thus improves on an already capable backbone in both contextual forecasting and temporal reasoning.

The individual experts reveal complementary strengths: native forecasting gives the best specialist CiK score (0.123), while analysis leads on TSE (78.418\%).
TeeMoE composition further reduces RCRPS by 6.6\% relative to the native expert, improves TSE accuracy by 0.134 percentage points over the analysis expert, and slightly improves the aggregation expert's GIFT mean rank.

Despite their different objectives and adapter capacities, the independently trained experts retain their strengths when composed within one backbone.
Task coverage can therefore be broadened without requiring a common training recipe.
The next comparison examines alternative adapter combinations and contrasts separate specialization with joint training.

\FloatBarrier
\subsubsection{Expert combination}

To understand how adapter combination and training affect the balance across capabilities, we compare TeeMoE with three composition controls and a jointly trained model.
Top-1 adapter routing activates the expert with the highest controller weight, assigning it unit weight and testing whether simultaneous adapter contributions provide a meaningful advantage.
Equal-weight and full-strength composition fix each adapter's coefficient to $1/3$ and $1$, respectively, retaining learned output selection.
Joint training uses one shared adapter on the three training populations, followed by a separate numerical-versus-text selector.
Since the specialists have different objectives and schedules, no unique joint counterpart exists; we prioritize capacity and stability through a common recipe that preserves their training exposure (Appendix~\ref{app:joint-comparison-scope}).

\begin{table}[htbp]
\centering
\caption{Composition controls and joint training: improvements over the pre-adaptation baseline (Table~\ref{tab:individual-experts}). Composition controls use the same frozen specialists; joint training uses one shared adapter and a separate output selector.}
\label{tab:expert-combination}
\includegraphics[width=\linewidth]{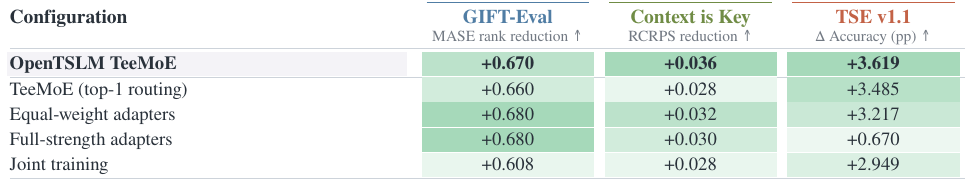}
\end{table}

\FloatBarrier

TeeMoE maintains comparable performance under top-1 routing, albeit with a noticeably smaller CiK improvement than soft composition and slightly worse GIFT and TimeSeriesExam performance.
TeeMoE's capability preservation therefore mostly extends across both soft composition and single-expert selection, with interactions among experts contributing only small additional predictive gains.
Equal weights slightly improve TeeMoE's GIFT rank, worsen CiK, and reduce TSE accuracy by 0.402 percentage points; full-strength adapters improve GIFT rank but worsen CiK and reduce TSE accuracy by 2.949 percentage points.
Thus, learned weighting offers a better balance than any tested fixed mixture, although a single selected specialist captures a good amount of that benefit.

Notably, TeeMoE outperforms the joint baseline on all three tasks: it achieves a larger CiK reduction (0.036 versus 0.028), a larger GIFT rank reduction (0.670 versus 0.608), and a 0.670-percentage-point gain in TSE accuracy.

\paragraph{Training and revision cost.}
Separate specialization offers practical benefits throughout model development.
Joint updates synchronize task-specific computations across GPUs, so differences in sequence lengths and loss calculations introduce coordination overhead.
Separate specialization costs 30.756 H100 GPU-hours, compared with 58.657 for our joint recipe (Table~\ref{tab:training-revision-cost}).
Appendix~\ref{app:modular-compute} details the training-cost accounting.
Another benefit is updating one capability while keeping the other experts fixed; even if joint training matched the cost of separate specialization, revising one expert and the controller would remain cheaper than a full joint retrain.

\begin{table}[htbp]
\centering
\caption{Training and expert-plus-controller revision costs in H100 GPU-hours, excluding shared numerical preparation.}
\label{tab:training-revision-cost}
\includegraphics[width=\linewidth]{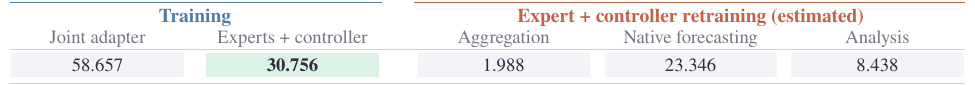}
\end{table}

\FloatBarrier
\subsubsection{Numerical ensembling and forecast refinement}

To study pooling, weighting, and refinement, we compare equal-weight eight- and thirteen-model pools, XGBoost-weighted pooling, Toto-FnF, and their fitted blend.
The blend supplies the aggregation expert's reference forecast.

\begin{table}[htbp]
\centering
\caption{Numerical ensembling and refinement, ordered by GIFT mean rank (best first). Numerical baselines use history only; TeeMoE also receives context on CiK. Protocols are in Appendix~\ref{app:capability-ablations}.}
\label{tab:numerical-baselines}
\includegraphics[width=\linewidth]{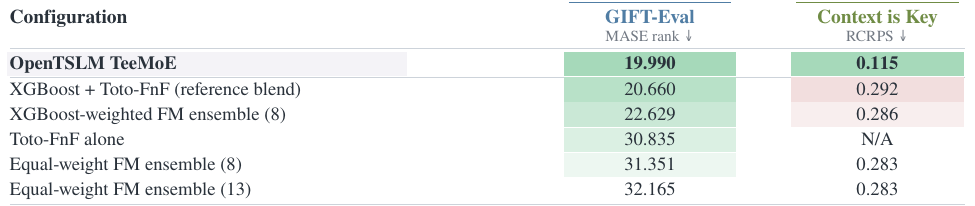}
\end{table}

On GIFT, learned weighting improves substantially over uniform pooling, whereas adding more models with equal weights does not help.
Note that refinement further improves the already strong blend, and a smaller rank gain does not imply a smaller contribution to forecast quality.
Because rank measures relative ordering rather than error magnitude, these gains cannot directly establish which component contributes more.

On CiK, the history-only ensembles score 0.283--0.292 RCRPS, well behind TeeMoE's 0.115.
This gap highlights the value of context-conditioned forecasting beyond numerical ensembling.

\section{Discussion and conclusion}

\method{} brings numerical forecasting, context-conditioned prediction, and temporal reasoning into one model, ranking among the top three systems on all three benchmarks.
The comparisons clarify how this breadth is achieved and where modularity is useful.

\paragraph{Preservation is the clearest benefit.}
Experts trained independently with different objectives retain their strengths when composed within one generalist.
Learned composition further improves contextual forecasting, with smaller gains on GIFT and TSE.
Although the limited predictive gains from routing are unsurprising, these results demonstrate that modular composition can produce a strong generalist model.

\paragraph{Why retain separate specialization?}
Apart from predictive performance, separate specialization offers practical advantages in training and revising the generalist.
Joint training must reconcile the capabilities' different objectives and optimization settings in one shared adapter; separate specialization lets each retain its own recipe.
In our comparison, this separation also lowers training cost (Table~\ref{tab:training-revision-cost}).
It allows an expert and the controller to be retrained while leaving the other experts fixed, so developing one capability need not require training all three again.
These are reasons to use a modular design even when composition preserves, rather than substantially exceeds, specialist accuracy.

\paragraph{Scope and next steps.}
The study also points to concrete next steps.
We prioritize GIFT-Eval, Context is Key, and TimeSeriesExam because they provide established evaluation protocols and a broad set of published specialist comparisons across our three capabilities.
These benchmarks assess capabilities separately, allowing us to evaluate whether a generalist preserves the strengths of specialized systems.
Extending this evaluation to workflows that alternate between forecasting and reasoning---for example, interpreting a forecast and revising it in response to additional context---is a natural next step.
Extending the study beyond one backbone family would also clarify how these findings carry over to other models and training recipes.

Taken together, the results support a practical route to generalist time-series modeling: train each capability with the objectives and resources it needs, then bring them together without giving up their strengths or the flexibility to develop them independently.

\section*{Reproducibility Statement}
The codebase is available on \href{https://github.com/OpenTSLM/OpenTSLM-TeeMoE}{GitHub}, and model checkpoints are available on \href{https://huggingface.co/OpenTSLM/TeeMoE}{Hugging Face}.
The codebase provides source preparation, expert training, composition, ablations, and comparison-model evaluation workflows. Appendix~\ref{app:architecture} records the detailed training recipes; Appendices~\ref{app:leaderboard-audit} and~\ref{app:baseline-evaluation} describe evaluation protocols.

\section*{Ethics Statement}
Forecasts and temporal interpretations can inform consequential decisions in healthcare, finance, and public infrastructure.
Benchmark accuracy does not establish fitness for autonomous decisions in those settings.
Users should validate the model on the intended domain and retain appropriate human oversight.
Appendix~\ref{app:impacts} discusses potential impacts and deployment considerations.

\section*{AI Use Statement}
\label{app:ai-use}

In this work, we used generative AI tools to help develop the conceptual framework, propose or refine hypotheses, design or provide feedback on research methodology and experiments, implement methods, and clean and reformat datasets.
We have not used generative AI tools for translation, formulating mathematical claims, qualitative or thematic data analysis, or directly generating dataset content, though AI-assisted code was used to implement the rule-based data augmentation described in Appendix~\ref{app:native-recipe}.
Tasks involving proving mathematical claims or writing proofs are not applicable to this work.
Additionally, we used generative AI tools to identify relevant literature, draft and edit parts of the paper, and create or modify figures.
AI tools assisted with monitoring experiments, summarizing results, and interpreting comparisons under author-defined criteria. The authors made the research decisions.
The authors directed the research and retain responsibility for verification of the AI-assisted work and the final content, including text, claims, and artifacts produced with the aid of generative AI.

\clearpage
\appendix
\section{Architecture and training details}
\label{app:architecture}

We first specify the request interface and numerical computation, then give the training recipes. Appendix~\ref{app:capability-ablations} describes the ablation protocols; Appendices~\ref{app:leaderboard-audit}, \ref{app:shared-prompt}, and \ref{app:baseline-evaluation} give evaluation protocols, exact prompts, and evaluations of released comparison models.

\subsection{Shared language interface}
\label{app:interface}

All textual requests use one static system instruction covering aggregation, native forecasting, and analysis.
Four user-message formats distinguish forecasting without context, forecasting with context, multiple-choice analysis, and free-form analysis.
These formats contain the example's history, timestamps where provided, context, question, and options as appropriate.
The same format is used for training and evaluation examples of the corresponding kind.
Appendix~\ref{app:shared-prompt} reproduces the shared instruction and four user-message templates.

Native forecasting and analysis share the language-model output head.
Aggregation uses a numerical encoder and forecast decoder.

\subsection{Controller and output selection}
\label{app:output-selection}

The controller produces one three-way softmax mixture per request, shared across adapted layers and held fixed for the response.
Let $h_{\mathrm{text}}$ and $h_{\mathrm{full}}$ be the final adapter-disabled hidden states for the task text and the full observed request, respectively; both include the shared system instruction.
The controller is
\begin{equation}
\begin{aligned}
 h&=(1-\alpha)h_{\mathrm{text}}+\alpha h_{\mathrm{full}},
 &\pi&=\operatorname{softmax}(Ah+b),\\
 \alpha&\in[0,1]\quad\text{(learnable)}.
\end{aligned}
\label{eq:controller-input}
\end{equation}
An aggregation weight strictly greater than $0.5$ selects the forecast decoder; otherwise generation uses language-model tokens.
Appendix~\ref{app:routing-example} illustrates the two input views for a contextual forecasting request.

\subsection{Numerical aggregation interface}
\label{app:numerical-interface}

\paragraph{Reference forecast.}
The aggregation expert first constructs a forecast distribution from frozen forecasting models, then uses the shared language backbone to refine it.
An XGBoost regressor---an ensemble of learned decision trees---predicts rank scores that determine weights for eight selected forecasting models using features of the observed series and their predictions.
Their combined forecast is blended with Toto-FnF to form the reference forecast that the numerical encoder presents to the backbone alongside the individual candidates.

\paragraph{Candidate forecasting models.}
The eight models combined by XGBoost share five predictors with the ten-member Toto-FnF ensemble. Their union supplies the editor's 13 candidates:
\begin{itemize}
  \item XGBoost combines TiRex 2~\citep{tirex2}, Toto 2.0 (2.5B)~\citep{toto2}, Chronos-2~\citep{chronos2}, Timer-S1~\citep{timers1}, TimesFM 2.5~\citep{timesfm25}, Moirai 2~\citep{moirai2}, FlowState~\citep{flowstate}, and PatchTST-FM~\citep{patchtstfm}.
  \item Including Toto-FnF contributes five additional models: TiRex 1.1~\citep{tirex} and the 4M, 22M, 313M, and 1B Toto 2.0 checkpoints~\citep{toto2}.
  \item Toto-FnF's full ten-member roster comprises Chronos-2, TimesFM 2.5, FlowState, TiRex 1.1, PatchTST-FM, and the five Toto 2.0 sizes (4M, 22M, 313M, 1B, and 2.5B)~\citep{totofnf}.
\end{itemize}

\paragraph{Numerical encoder.}
The encoder, shown as the numerical connector in Figure~\ref{fig:system}, represents the reference quantile forecast $q_0$ and 13 candidate distributions as continuous input tokens for the language backbone. Each distribution uses nine quantile levels, $0.1,0.2,\ldots,0.9$.
For a horizon of $H$ steps, it represents the distributions at eight rounded, evenly spaced indices from $0$ to $H-1$.
At each selected step, a 53-value feature vector describes the reference forecast's quantiles, the candidates' medians, widths and asymmetries, and the step's position and scale.
Quantiles and medians are expressed relative to the reference median; forecast offsets and widths are divided by the scale $s_t$ defined below.
The 53 coordinates comprise nine reference-quantile offsets, thirteen candidate-median offsets, thirteen widths $(q_{e,t,0.9}-q_{e,t,0.1})/s_t$, thirteen asymmetries $(q_{e,t,0.9}+q_{e,t,0.1}-2m_{e,t})/s_t$, four position coordinates, and one scale coordinate.
For sampled position $j\in\{0,\ldots,7\}$, the position coordinates are $p_j=j/7$, $\sin(2\pi p_j)$, $\cos(2\pi p_j)$, and $\log 8$; the scale coordinate is $\log(1+s_t)$.
All descriptor coordinates are clipped to $[-20,20]$.

The encoder also represents how each candidate differs from the reference across all nine quantiles, normalizing these differences by $s_t$.
Their means and standard deviations, computed over the eight XGBoost candidates and over all thirteen candidates, summarize agreement across models.
Learned linear projections and layer normalization combine these differences, group summaries, and the 53-value feature vector into one 5,120-dimensional embedding per selected step.
A learned position embedding is added to each.

The eight resulting forecast embeddings appear three times in one 24-token sequence: first in reverse order, then twice in forward order.
The first two blocks supply forecast evidence; the final eight hidden states feed the forecast decoder.
This ordering presents the forecast grid in both directions before the output block, so every output position can attend to evidence from the full horizon under causal attention.
All three blocks are processed in one backbone pass.

\paragraph{Forecast decoder.}
A linear head produces 14 outputs at each of the final eight positions: one controls the correction's magnitude and sign, while the other 13 weight candidate medians to form a comparison forecast.
The 14 raw head channels are linearly interpolated across the full horizon before the softmax and bounded-correction nonlinearities are applied.
Interpolation uses the eight uniformly spaced decoder positions with aligned endpoints. When rounded input indices repeat for horizons shorter than eight, all eight decoder positions are retained.
The head is initialized to zero, so the initial forecast equals the reference.
The correction rule below defines the resulting forecast; Appendix~\ref{app:aggregation-recipe} specifies how it is fitted.

\paragraph{Editor correction rule.}
Let $q_{0,t,\tau}$ denote the reference forecast at time $t$ and quantile level $\tau$, and let $m_{0,t}$ and $m_{e,t}$ be the reference and candidate medians.
The output head emits a scalar $a_t$ and 13 candidate logits $b_{e,t}$.
After interpolating these outputs to the full horizon, the normalized correction is
\begin{equation}
  d_t=\frac{1}{2}\tanh(a_t)
  \tanh\!\left(
    \frac{\sum_e [\operatorname{softmax}(b_t)]_e\,m_{e,t}-m_{0,t}}{s_t}
  \right).
\end{equation}
Here $s_t>0$ is the normalization scale defined below.
The inference rule reduces the correction when candidate disagreement is large relative to reference uncertainty:
\begin{equation}
  q_{t,\tau}=q_{0,t,\tau}+g_t s_t d_t,
  \qquad g_t=\frac{v_t}{v_t+r_t^2}.
\end{equation}
Candidate disagreement, reference uncertainty, and the normalization scale are
\[
\begin{aligned}
 r_t&=\operatorname{median}_{e}|m_{e,t}-m_{0,t}|,\\
 v_t&=\max(q_{0,t,0.7}-q_{0,t,0.3},\max(10^{-4}|m_{0,t}|,10^{-3}))^2,\\
 s_t&=\max(r_t,\delta_t,10^{-4}|m_{0,t}|,10^{-3}),
\end{aligned}
\]
where $\delta_t$ is the lower median of the eight nonnegative differences between adjacent reference quantiles.
The quantities $r_t$, $s_t$, and $\sqrt{v_t}$ are in signal units; $d_t$ and $g_t$ are dimensionless.
The same shift applies to every quantile at a given time step, preserving quantile order and within-step width.
The disagreement gate is applied at inference; training supervises the ungated normalized correction $d_t$.

\subsection{Shared text-training objective}

Native forecasting and analysis use teacher-forced response-token prediction with KL regularization to the frozen base model.
Their common objective is
\begin{equation}
\begin{aligned}
  \mathcal L_{\mathrm{text}}
    &=\frac{1}{N}\sum_{i\in\mathcal B}\sum_{t\in\mathcal T_i}\ell_{i,t},
    \qquad N=\sum_{i\in\mathcal B}|\mathcal T_i|,\\
  \ell_{i,t}
    &=-\log p_\theta(y_{i,t}\mid x_i,y_{i,<t})
      +\beta_{\mathrm f}D_{\mathrm{KL}}(p_0\Vert p_\theta)
      +\beta_{\mathrm r}D_{\mathrm{KL}}(p_\theta\Vert p_0).
\end{aligned}
\label{eq:text-training}
\end{equation}
Here $\mathcal B$ is the global batch, $\mathcal T_i$ contains example $i$'s response-token positions, and $x_i$ is its input prompt.
In both KL terms, the frozen distribution $p_0$ and adapted distribution $p_\theta$ are conditioned on the same prefix $(x_i,y_{i,<t})$; all terms are averaged over the same response tokens.
Native forecasting uses $(\beta_{\mathrm f},\beta_{\mathrm r})=(0.5,0)$, while analysis uses $(0.05,0.05)$.
Appendices~\ref{app:native-recipe} and~\ref{app:analysis-recipe} specify their data and optimization settings.

\subsection{Aggregation data and hyperparameters}
\label{app:aggregation-recipe}

Training has three stages: fit XGBoost to combine eight selected forecasting models, learn a constant weight that blends its forecast with Toto-FnF, and train the aggregation expert to refine the resulting reference forecast.
All forecasting models, including the released Toto-FnF ensemble, are frozen.

Prior exact-window and forecast-interval audits found no overlap with GIFT-Eval evaluation targets.
All 13 constituent forecasters and the Toto-FnF ensemble are also listed as having no test-data leakage on the GIFT-Eval leaderboard~\citep{gifteval_leaderboard}.

\paragraph{Cross-fitting the editor's reference forecasts.}
The editor learns to correct a forecast supplied by the fitted numerical ensemble.
To train it on forecasts for series unseen by the XGBoost model, we divide the 576,920-window ensemble-training population into ten folds by source series.
For each fold, we fit an XGBoost model and a Toto-FnF blend weight using the other nine folds, then construct reference forecasts for the held-out GIFT training windows.
Repeating this process across the ten folds supplies the editor's pool of 296,161 eligible windows. Each reference is paired with its corresponding observed future as the correction target.
Cross-fitting applies to the XGBoost model and scalar blend; the pretrained forecasting models are reused unchanged across folds.
At inference, a single XGBoost ensemble and blend weight fitted on the full population supply the reference forecast.
The following recipes specify the fitting rules and editor sampling.

\begin{recipebox}{Stage 1: XGBoost ensemble}
\textbf{Hyperparameters.}
1,200 boosting rounds; maximum depth 10; learning rate 0.04; minimum child weight 12; row and tree-column sampling fractions 0.9; $L_2$ coefficient 2; $L_1$ coefficient 0.05; maximum histogram bins 256.

\smallskip
\textbf{Features and objective.}
The same regressor scores all eight models: 195 history/forecast coordinates expand to 1,433 expert-conditioned features. Its target is $\log(1+R_{i,e})$, where $R_{i,e}\in\{0,\ldots,7\}$ ranks candidate $e$ by empirical CRPS within window $i$, with zero denoting the lowest risk. Squared-error fitting uses unit example weights.
Predicted log-ranks are centered and standardized within each row with a standard-deviation floor of 0.15. Writing these standardized scores as $z_{i,e}$, the allocation and pooled distribution are
\[
 w_{i,e}=\operatorname{softmax}(-z_i/1.35)_e,
 \qquad F_{\mathrm{XGB},i}(u)=\sum_{e=1}^{8}w_{i,e}F_{i,e}(u).
\]
Here $F_{i,e}$ is the discrete CDF placing equal mass on candidate $e$'s nine supplied quantile values at each forecast step. We sort the pooled values, accumulate their weighted masses, and select the first value whose cumulative mass reaches each requested quantile level. This construction has bounded support at the supplied values. The feature schema and empirical-CRPS construction are provided in the reproduction code.

\recipedivider
The 576,920-window training population combines the GIFT training split~\citep{gifteval}, observability windows from the Benchmark of Observability Metrics (BOOM)~\citep{boom_data}, and forecasting collections and retail sources from RMISC~\citep{rmisc_data}.
The component drawn from the Large-scale Open Time Series Archive (LOTSA)~\citep{moirai} balances seven domains, while the 1G subset of the Unified Time Series Dataset (UTSD)~\citep{timer} supplies the listed supplementary series.

\begin{recipedatalist}
  \recipedataitem{GIFT training split}{300,000}
  \recipedataitem{BOOM}{100,000}
  \recipedataitem{RMISC}{58,000}
  \begin{recipedatalist}
    \recipedataitem{Multivariate forecasting collection}{50,000}
    \recipedataitem{Additional retail datasets}{8,000}
    \begin{recipedatalist}
      \recipedataitem{Dominick}{4,000}
      \recipedataitem{Rossmann (weekly)}{4,000}
    \end{recipedatalist}
  \end{recipedatalist}
  \recipedataitem{LOTSA-derived seven-domain corpus}{100,000}
  \begin{recipedatalist}
    \recipedataitem{Economics and finance}{14,286}
    \recipedataitem{Energy}{14,286}
    \recipedataitem{Healthcare}{14,286}
    \recipedataitem{Nature}{14,286}
    \recipedataitem{Sales}{14,286}
    \recipedataitem{Transport}{14,285}
    \recipedataitem{Web and cloud operations}{14,285}
  \end{recipedatalist}
  \recipedataitem{UTSD-1G}{18,920}
  \begin{recipedatalist}
    \recipedataitem{Health: IEEEPPG}{4,000}
    \recipedataitem{Health: SelfRegulationSCP1}{4,000}
    \recipedataitem{Health: SelfRegulationSCP2}{4,000}
    \recipedataitem{Health: TDBrain}{4,000}
    \recipedataitem{Health: AtrialFibrillation}{384}
    \recipedataitem{Nature: Worms}{1,648}
    \recipedataitem{IoT: baian}{696}
    \recipedataitem{Environment: BenzeneConcentration}{144}
    \recipedataitem{Environment: AustraliaRainfall}{48}
  \end{recipedatalist}
\end{recipedatalist}
\noindent\textbf{Total windows}\hfill\textbf{576,920}
\smallskip

Each cross-fitting XGBoost model uses the nine non-held-out folds of this population; the deployment model uses the full population. Exact source coordinates, identities, extraction rules, and retained-row selections are specified by the code's source catalogs and manifests.
\end{recipebox}

\begin{recipebox}{Stage 2: Toto-FnF blend weight}
\textbf{Fitting rule.}
For each informative example, find the weight in $[0,1]$ that minimizes the weighted absolute error of the blended XGBoost ensemble and Toto-FnF point forecasts at eight horizon positions. Take the arithmetic mean of these per-example weights. The full-data estimate is $\alpha\approx0.386$ for Toto-FnF.

\smallskip
We exclude examples with an unavailable forecast or no valid difference between the two forecasts. Each cross-fitting fold estimates its own scalar from the other nine folds using the same rule. At inference, corresponding quantiles are blended with the fixed learned weight:
\[
 q_{0,t,\tau}=(1-\alpha)q_{\mathrm{XGB},t,\tau}
              +\alpha q_{\mathrm{FnF},t,\tau}.
\]
This second-stage quantile blend is distinct from the eight-model CDF pooling in Stage 1. We use the XGBoost ensemble alone if Toto-FnF is unavailable.

\recipedivider
\begin{tabularx}{\linewidth}{@{}Xr@{}}
\recipedatarow{Candidate windows (the same sources as Stage 1)}{576,920}
\recipedatarow{Informative rows used in the full-data scalar fit}{465,182}
\end{tabularx}
\end{recipebox}

\begin{recipebox}{Stage 3: Foundation model aggregation expert}
\textbf{Model and trainable parameters.}
Qwen3.6-27B with LoRA rank 4, alpha 8, and dropout 0.
LoRA adapts the attention, recurrent-mixing, and feed-forward projections while the backbone remains frozen.
The numerical encoder and 14-channel output head are also trained.

\smallskip
\textbf{Objective.}
The normalized target correction is $d_t^*=\operatorname{clip}((y_t-m_{0,t})/s_t,-0.5,0.5)$, with per-row loss
\begin{equation}
  \mathcal L_i=\sum_t \widetilde w_{i,t}
  \left[\operatorname{SmoothL1}_{0.05}(d_{i,t},d_{i,t}^*)+0.2d_{i,t}^2\right],
  \qquad \sum_t\widetilde w_{i,t}=1.
\end{equation}
Here $y_t$ is the observed future target, and $t$ ranges over the eight cached horizon positions. Each original horizon step is assigned to its nearest sampled position, with ties assigned to the first such position. The resulting counts, divided by the horizon length, define $\widetilde w_{i,t}$; row losses are averaged across the batch.
The loss fits $d_t$, and the squared-edit term regularizes this ungated correction.

\smallskip
\textbf{Optimization.}
AdamW uses constant learning rate $3\times10^{-5}$, weight decay 0, moments $(0.9,0.999)$, and gradient-norm clipping 1.
Training comprises 128 updates at global batch size 32, for 4,096 sampled training examples.
The backbone computes in BF16; trainable parameters and optimizer state use FP32.
\recipedivider
\begin{recipedatalist}
  \recipedataitem{GIFT training split: eligible editor windows}{296,161}
\end{recipedatalist}
\smallskip
These windows form the editor's sampling pool within Stage 1's GIFT training population.
Each example pairs 13 candidate distributions and a cross-fitted reference forecast with future observations at eight horizon positions.

\smallskip
We sample with replacement.
\end{recipebox}

\subsection{Native forecasting data and hyperparameters}
\label{app:native-recipe}

The native expert learns to generate future values from the observed history and any supplied context through the language-model output head.

The mixture combines published synthetic pairs from TADiff~\citep{tadiff_data}, National Weather Service (NWS) forecast discussions~\citep{nws_archive}, contextual series from CAF-7M~\citep{caf_data}, and author-released synthetic time series~\citep{pierrot_pinson_data}.
It also includes windows from the Large-scale Open Time Series Archive (LOTSA)~\citep{moirai} with imposed bounds.
The selected TADiff component contains 3,600 complete pairs: the two members share an observed history and retain their respective author-supplied future captions and trajectories.
Pierrot--Pinson examples provide the released support schedule as context; LOTSA bounds are introduced by the augmentation described below.

\begin{recipebox}{Native forecasting adapter}
\textbf{Model and trainable parameters.}
Qwen3.6-27B with LoRA rank 32, alpha 64, and dropout 0.
LoRA adapts the attention, recurrent-mixing, and feed-forward projections.
The backbone and vocabulary head remain frozen.

\smallskip
\textbf{Objective.}
Response-token cross-entropy with forward KL regularization to the frozen base model: $(\beta_{\mathrm f},\beta_{\mathrm r})=(0.5,0)$ in Equation~\ref{eq:text-training}.

\smallskip
\textbf{Optimization.}
AdamW uses peak learning rate $10^{-5}$, weight decay $0.01$, and gradient-norm clipping 1.
Training comprises one epoch at global batch size 8.
The learning rate follows a 2,500-update cosine schedule with 125 warmup updates and a final rate of $10^{-6}$.
The backbone computes in BF16; trainable parameters and optimizer state use FP32.

\recipedivider
\begin{recipedatalist}
  \recipedataitem{TADiff}{7,200}
  \recipedataitem{NWS}{2,800}
  \recipedataitem{CAF}{2,800}
  \recipedataitem{Pierrot--Pinson}{3,600}
  \recipedataitem{LOTSA (augmented)}{3,600}
\end{recipedatalist}
\noindent\textbf{Total examples}\hfill\textbf{20,000}

\smallskip
The CAF source pool excludes 55 reviewed examples whose stated bounds contradict their numerical targets.
We augment real LOTSA windows to teach the model to prioritize explicit context over extrapolation from past observations: we leave the history unchanged, clip future values to upper and/or lower bounds, and state those bounds in the context.
All examples use the shared system instruction and the ordinary or contextual forecasting user format in Appendix~\ref{app:shared-prompt}. Source-window coordinates, retained-pair membership, and preparation commands are supplied with the reproduction code.

\end{recipebox}

\subsection{Analysis data and hyperparameters}
\label{app:analysis-recipe}

The analysis expert maps observed signals and questions to categorical or free-form answers through the language-model head.
Its training mixture contains activity/anomaly questions from Time-MQA~\citep{timemqa_data}, questions from ChengsenWang/TSQA~\citep{chattime_data} and HiTSR~\citep{llatisa_data}, and synthetic examples from the ChatTS authors' published generator~\citep{chatts}.
These examples cover activity recognition, anomalies, signal comparisons, numerical readouts, and temporal descriptions.
Source answers and their requested formats are retained.

\begin{recipebox}{Analysis adapter}
\textbf{Model and trainable parameters.}
Qwen3.6-27B with LoRA rank 16, alpha 32, and dropout 0.
LoRA adapts the attention, recurrent-mixing, and feed-forward projections.
The backbone and vocabulary head remain frozen.

\smallskip
\textbf{Objective.}
Response-token cross-entropy with forward and reverse KL regularization to the frozen base model: $(\beta_{\mathrm f},\beta_{\mathrm r})=(0.05,0.05)$ in Equation~\ref{eq:text-training}.

\smallskip
\textbf{Optimization.}
AdamW uses peak learning rate $10^{-5}$, weight decay 0, and gradient-norm clipping 1.
Training comprises one epoch at global batch size 8.
The learning rate follows a 1,500-update cosine schedule with 75 warmup updates and a final rate of $10^{-6}$.
The backbone computes in BF16; trainable parameters and optimizer state use FP32.

\recipedivider
\begin{recipedatalist}
  \recipedataitem{Time-MQA}{4,000}
  \begin{recipedatalist}
    \recipedataitem{Activity classification}{2,000}
    \recipedataitem{Anomaly detection}{2,000}
  \end{recipedatalist}
  \recipedataitem{ChengsenWang/TSQA}{3,000}
  \recipedataitem{HiTSR}{2,000}
  \begin{recipedatalist}
    \recipedataitem{Level 1}{1,087}
    \recipedataitem{Level 2}{113}
    \recipedataitem{Period comparisons}{615}
    \recipedataitem{Spike comparisons}{185}
  \end{recipedatalist}
  \recipedataitem{ChatTS author-generated UTS examples}{3,000}
  \begin{recipedatalist}
    \recipedataitem{Descriptions}{449}
    \recipedataitem{Numerical readouts}{448}
    \recipedataitem{Structured attribute descriptions}{448}
    \recipedataitem{Local fluctuations}{1,655}
  \end{recipedatalist}
\end{recipedatalist}
\noindent\textbf{Total examples}\hfill\textbf{12,000}

\smallskip
The ChatTS examples cover 449 distinct signals; we use the authors' published generator without modification.
Training and evaluation use the same input preparation and analysis format (Appendix~\ref{app:analysis-prompts}), including any source-supplied definitions or clarifications.
\end{recipebox}

\subsection{Composition data and hyperparameters}
\label{app:composition-recipe}

Composition learns request-conditioned weights over the three completed experts, keeping the numerical ensemble, adapters, encoder, and output heads frozen. Controller examples are sampled from the expert training populations listed below.

\begin{recipebox}{LoRA mixture-of-experts controller}
\textbf{Model and trainable parameters.}
The backbone, all three adapters, numerical encoder, and numerical output head are frozen.
A linear map from the 5,120-dimensional request representation to three logits and the scalar $\alpha$ in Equation~\ref{eq:controller-input} are trained, with a temperature-one softmax.
The map's weights and bias are initialized to zero, giving a uniform mixture. The resulting expert weights are shared across adapted layers.

\smallskip
\textbf{Objective.}
Text examples use response-token cross-entropy, averaged over that capability's response tokens. Numerical examples use the ungated normalized-correction objective in Appendix~\ref{app:aggregation-recipe}, with SmoothL1 transition width 0.05 and no squared-edit penalty, averaged over weighted rows.
For capability $c$, let $\mathcal L_c$ denote this prediction loss and $\mathcal H_c$ the example-averaged output-format loss. Each update minimizes
\[
  \mathcal L_{\mathrm{ctrl}}=\frac{1}{3}\sum_c
             \bigl(\mathcal L_c+\mathcal H_c\bigr).
\]
The format loss is $-\log\pi_{\mathrm{agg}}$ for numerical examples and $-\log(1-\pi_{\mathrm{agg}})$ for text examples. Its coefficient is 1 within each capability mean.
This supervises the numerical-versus-text output choice without prescribing native-versus-analysis weights or the full three-expert mixture.

\smallskip
\textbf{Optimization.}
AdamW uses constant learning rates $\eta_{A,b}=10^{-4}$ and $\eta_{\alpha}=0.05$, with weight decay 0 and $\alpha\leftarrow\operatorname{clip}_{[0,1]}(\alpha)$ after each update.
Training comprises one epoch at global batch size 24. Full batches contain 8 examples from each capability; the final batch contains the remaining examples.
The backbone computes in BF16; trainable parameters and optimizer state use FP32.

\recipedivider
\begin{recipedatalist}
  \recipedataitem{Aggregation training windows}{334}
  \recipedataitem{Native forecasting training examples}{333}
  \recipedataitem{Analysis training examples}{333}
\end{recipedatalist}
\noindent\textbf{Total examples}\hfill\textbf{1,000}

\smallskip
Examples are sampled without replacement from the three expert-training populations described above.
For each request, the controller computes the mixture once and holds it fixed throughout that example's numerical editing or text generation, in both training and inference.
Exact selected-row identities and training plans are retained in the reproduction metadata.
\end{recipebox}

\section{Ablation protocols}
\label{app:capability-ablations}

This appendix specifies the expert and ensemble controls in Tables~\ref{tab:individual-experts}--\ref{tab:numerical-baselines}.
Benchmark protocols and comparison sources are described in Appendix~\ref{app:leaderboard-audit}.

\subsection{References for ablation deltas}
\label{app:ablation-deltas}

Tables~\ref{tab:individual-experts} and~\ref{tab:expert-combination} use fixed task-specific baselines. Let $r$ denote mean MASE rank over the 97 GIFT cells, $c$ the full CiK RCRPS, and $a$ TSE accuracy as a fraction.
We report $r_0-r$ for GIFT, $c_0-c$ for CiK, and $100(a-a_0)$ percentage points for TSE.
The pre-edit numerical ensemble defines $r_0$; the unadapted Qwen3.6-27B evaluations define $c_0$ and $a_0$.
The displayed reference scores are 20.660, 0.151, and 74.933\%; calculations use the unrounded values.
The ``Pre-adaptation baseline'' row combines the fitted numerical ensemble on GIFT with unadapted Qwen on CiK and TSE.
All GIFT ranks use the same pinned 133-submission comparison roster, with each evaluated configuration inserted separately.

\subsection{Individual experts and composition controls}

For CiK and TimeSeriesExam, every individual-expert condition uses the full textual request, shared instruction, language-model output head, and benchmark scorer. It activates only the named frozen adapter, without computing controller weights.
The aggregation expert therefore uses its LoRA with the language-model head on CiK and TimeSeriesExam, and its complete numerical interface on GIFT.
The matched analysis specialist uses a one-hot mixture in the composition runtime; the matched CiK text comparisons use vLLM with the same sampling settings as composition.
The native and analysis experts' GIFT entries follow the OOT protocol in Appendix~\ref{app:gift-runtime}.

Soft composition uses the learned mixture and selects the forecast decoder only when $\pi_{\mathrm{agg}}>0.5$ (Appendix~\ref{app:output-selection}).
Top-1 adapter routing replaces that mixture by a one-hot vector for its largest-weight expert, assigning it unit weight, then uses the selected expert's output head.

\paragraph{Unadapted backbone.}
\label{app:raw-backbone}
The Qwen3.6-27B baseline uses the same pinned pretrained backbone without any trained adapter.
We evaluate this baseline on all 355 CiK instances with 25 trajectories per instance and all 746 TSE v1.1 questions.
CiK uses the shared forecasting prompt and syntax-constrained numerical generation.
The TSE control uses the same complete inputs as TeeMoE for all 746 questions, with greedy 64-token decoding.

\paragraph{Equal-weight adapters.}
Each frozen adapter receives coefficient $1/3$, retaining its saved LoRA scaling and the controller's learned output selection. The coefficients sum to one for every request.

\paragraph{Full-strength adapters.}
Each frozen adapter receives coefficient one, preserving its saved LoRA scaling. The controller selects the numerical or textual output path using its learned weights; that path then executes with all three adapters at full strength. All other parameters and evaluation settings are unchanged.

\subsection{Numerical ensemble controls}

On GIFT, each configuration is inserted separately into the same 133-submission leaderboard roster over the same 97 evaluation cells.
The equal-weight conditions pool either the core eight or all 13 candidate CDFs uniformly. XGBoost alone uses the fitted weights over the core eight; Toto-FnF alone uses its forecast directly. Their fitted blend is the numerical-ensemble-only condition, which omits the aggregation expert's refinement.

On CiK, the numerical ensembles receive observed history without textual context on all 355 instances, with no fitting on CiK.
The learned ensemble retains the XGBoost model and scalar blend fitted on GIFT training data and external corpora. Each score uses one sampling run with 25 trajectories drawn independently across forecast steps from inverse marginal CDFs.
For the 15 instances at second-level frequency unsupported by Toto-FnF, the fitted blend uses its XGBoost ensemble forecast; the equal-weight pools retain their respective candidate sets. Toto-FnF alone is reported as N/A because it cannot cover the full benchmark.

\subsection{Joint-training baseline}
\label{app:joint-comparison-scope}

Our comparison principle is to retain the specialists' data exposure and prediction interfaces while replacing separate adapters with shared parameters.
The joint baseline uses the same frozen backbone, adapted modules, numerical interfaces, training examples, and number of presentations per capability, with one rank-32 adapter serving all three capabilities.
It retains the numerical and textual output paths, prediction targets, and aggregation loss and edit penalties.

\paragraph{Why rank 32 rather than 52?}
We use rank 32 to share capacity without increasing the largest specialist's training-time adapter width.
A natural objection is that the specialist ranks sum to $4+32+16=52$, suggesting a shared rank-52 adapter.
That sum matches storage, but not how the capacity is trained: aggregation examples originally train a rank-4 adapter, native examples rank 32, and analysis examples rank 16; a shared rank-52 adapter makes all 52 components trainable by every task.
Nor must useful adaptation dimensions add across separately stored adapters: the original LoRA study finds overlapping leading directions across different ranks and competitive performance at low ranks~\citep[Section~7.2]{lora}.
Using rank 52 would therefore combine parameter sharing with an increase beyond every specialist's training-time rank, adding a capacity-expansion confound.
Rank 32 is the more direct comparison under our width-matching criterion: it retains the largest specialist's rank and scaling while allowing all tasks to train and reuse that capacity.

\paragraph{Learning rate and schedule.}
To preserve the established optimization scale as directly as possible, we use the textual specialists' $10^{-5}$ peak learning rate and cosine schedule.
These settings support the two capabilities supplying 32,000 of the 36,096 training presentations and provide our stability-oriented choice for the shared adapter.
Using the aggregation expert's $3\!\times\!10^{-5}$ rate globally would instead triple the peak rate for both textual capabilities, departing further from their specialist recipes.

\paragraph{Resulting hyperparameters.}
Table~\ref{tab:joint-hyperparameters} implements these choices; tuples follow aggregation, native forecasting, and analysis.
Median and majority rules retain the settings shared by the two textual specialists; weight decay uses the arithmetic mean rounded to one significant digit.

\begin{table}[htbp]
\centering
\small
\caption{Joint settings derived from the three specialist recipes. Tuples follow aggregation, native forecasting, and analysis; $\operatorname{round}_{1\mathrm{sf}}$ rounds to one significant digit.}
\label{tab:joint-hyperparameters}
\begin{tabular}{ll}
\toprule
Setting & Rule and joint value \\
\midrule
LoRA rank & $r_{\mathrm J}=32$ \\
LoRA scaling / dropout & $\alpha_{\mathrm J}=2r_{\mathrm J}=64$; dropout $0$, inherited \\
Global batch size & $B_{\mathrm J}=\operatorname{median}(32,8,8)=8$ \\
Peak learning rate & $\eta_{\mathrm J}=\min(3\!\times\!10^{-5},10^{-5},10^{-5})=10^{-5}$ \\
Weight decay & $\lambda_{\mathrm{wd},\mathrm J}=\operatorname{round}_{1\mathrm{sf}}((0+0.01+0)/3)=0.003$ \\
Schedule & $\operatorname{mode}(\text{constant},\text{cosine},\text{cosine})=\text{cosine}$ \\
Minimum/peak rate & $\rho_{\mathrm J}=\operatorname{median}(1,0.1,0.1)=0.1$ \\
\bottomrule
\end{tabular}
\end{table}

\paragraph{Predictor training.}
Training exposure is inherited as $N_c=U_cB_c$: 4,096 aggregation, 20,000 native, and 12,000 analysis presentations, giving $36{,}096/8=4{,}512$ updates.
Examples are globally shuffled into batches of eight.
Aggregation retains 4,072 distinct cached rows and 24 repeated presentations.
Each capability's mean loss is weighted by its fraction of examples in the batch, retaining response-token means for text and weighted-row means for aggregation; numerators and denominators are reduced globally across devices.
Aggregation retains Smooth L1 transition $0.05$, squared-edit coefficient $0.2$, and its correction targets and row/horizon weights.
All prediction-loss coefficients are one; there is no output-choice loss in this stage.

The adapter and numerical encoder/decoder are freshly initialized, with zero initial numerical correction and the same adapted modules as the specialists.
The backbone and language-model head remain frozen in BF16; trainable modules use FP32.
The adapter and numerical parameter groups follow one presentation clock: linear warmup over the first $\lceil0.05\times36{,}096\rceil=1{,}805$ presentations, then cosine decay to $10^{-6}$ at the final presentation.
We clip the combined gradient once per update; numerical parameters and their optimizer state advance only when aggregation examples are present.

\paragraph{Separate output selection.}
After freezing the joint predictor, we train a binary linear selector on the same 1,000 adapter-disabled instruction/context representations used for composition: 334 aggregation, 333 native, and 333 analysis examples.
For the 5,120-dimensional request representation, it predicts $p_{\mathrm{num}}=\sigma(v^\top h+b)$ and selects numerical output when $p_{\mathrm{num}}>0.5$, otherwise text.
Zero weights and bias $\log(1/2)$ initialize $p_{\mathrm{num}}=1/3$, matching an initially uniform three-expert controller.
Binary cross-entropy uses aggregation as the positive class and equally weights the three capability means, retaining TeeMoE's output-format supervision.
The selector inherits composition's AdamW settings: constant learning rate $10^{-4}$, zero weight decay, moments $(0.9,0.999)$, $\epsilon=10^{-8}$, and clipping norm one.
One pass gives $\lceil1000/24\rceil=42$ updates, with eight examples per capability in full batches and the remaining examples in the final batch.
Both output paths use the same shared adapter.

Full evaluation covers the same 97 GIFT cells, 355 CiK instances with 25 trajectories each, and 746 TSE questions.

\paragraph{Scope of the joint comparison.}
This configuration provides a reproducible shared-adapter comparison with matched training exposure and prediction interfaces.
A broader study of joint-training configurations is left to future work.

\section{Evaluation protocols and comparison sources}
\label{app:leaderboard-audit}

We exclude reports that use evaluation data for fitting without an established
held-out split, evaluate a different test population or supervision protocol,
or omit the benchmark's ranking metric. Tables~\ref{tab:cik-omissions}
and~\ref{tab:tse-omissions} document each exclusion and its source.
CiK and TSE comparison sources were last checked on September 14, 2026; reported scores
are taken directly from the cited sources.

\subsection{GIFT-Eval}

From the pinned official GIFT-Eval result tree~\citep{gifteval_results}, we retain 133 complete reference
submissions covering the same 97 evaluation cells (371,330 forecast rows). Table~\ref{tab:leaderboards}
adds our composed model and recomputes every method's per-cell
midrank over the resulting 134-entry population.  Consequently, the
reference-only ranks for STRIDE + Synapse and EXAONE Forecast Agent, 15.124
and 19.495, become 15.412 and 19.928 after our entry is included; our
corresponding rank is 19.990 and its position is third.
The reference tree was verified against the official repository on September 23, 2026.
The comparison includes individual forecasting models, forecasting ensembles such as STRIDE + Synapse, and agent-based systems such as EXAONE Forecast Agent.
\paragraph{Selected public comparison models.}
The lower blocks of Table~\ref{tab:leaderboards} include the same seven models
across benchmarks: TimesFM-3, Chronos-2, Toto-2.0-2.5B, TimeOmni-VL,
TS-Reasoner-7B, ChatTS-14B, and OpenTSLM SP Llama 3.2 1B~\citep{timeomnivl,tsreasoner,chatts,opentslm}.
The FM GIFT values use the same pinned joint roster as the upper block.
The notation n/a denotes an unsupported output interface: the numerical FMs
do not directly answer TSE questions.
Published ChatTS and TS-Reasoner accuracies on older TSE populations are not
substituted for v1.1 measurements. Local baseline evaluations are marked with
an asterisk and documented in Appendix~\ref{app:baseline-evaluation}.
OOT marks comparisons projected to exceed our compute budget
(Appendix~\ref{app:gift-runtime}).

\begingroup
\small
\setlength{\tabcolsep}{7pt}
\renewcommand{\arraystretch}{1.1}
\setlength{\LTcapwidth}{\textwidth}
\begin{longtable}{@{}lrrrr@{}}
\caption{TeeMoE on all 97 GIFT-Eval dataset--frequency--horizon categories (371,330 forecast windows). Ranks use the same 134-entry population as Table~\ref{tab:leaderboards}. Category errors are raw. Overall follows the official leaderboard: geometric means of seasonal-naive-normalized errors and mean ranks, equally weighting categories~\citep{gifteval_leaderboard}. S/M/L denote short/medium/long horizons. Lower is better throughout.}\label{tab:gift-all-categories}\\
\toprule
Category & MASE & MASE rank & CRPS & CRPS rank \\
\midrule
\endfirsthead
\multicolumn{5}{l}{Table~\thetable{} continued.}\\
\toprule
Category & MASE & MASE rank & CRPS & CRPS rank \\
\midrule
\endhead
\midrule
\multicolumn{5}{r}{Continued on next page.}\\
\endfoot
\bottomrule
\endlastfoot
m4\_yearly/A/S & 2.952 & 5.000 & 0.105 & 7.000 \\
m4\_quarterly/Q/S & 1.098 & 4.000 & 0.071 & 8.000 \\
m4\_monthly/M/S & 0.881 & 5.000 & 0.089 & 7.000 \\
m4\_weekly/W/S & 1.820 & 3.000 & 0.034 & 3.000 \\
m4\_daily/D/S & 3.165 & 18.000 & 0.021 & 12.000 \\
m4\_hourly/H/S & 0.644 & 10.000 & 0.018 & 8.000 \\
electricity/15T/S & 0.918 & 26.000 & 0.078 & 22.000 \\
electricity/15T/M & 0.791 & 11.000 & 0.071 & 20.000 \\
electricity/15T/L & 0.840 & 10.000 & 0.070 & 14.000 \\
electricity/H/S & 0.885 & 27.000 & 0.064 & 28.000 \\
electricity/H/M & 1.019 & 10.000 & 0.072 & 17.000 \\
electricity/H/L & 1.117 & 13.000 & 0.079 & 13.000 \\
electricity/D/S & 1.348 & 9.000 & 0.053 & 9.000 \\
electricity/W/S & 1.366 & 8.000 & 0.044 & 6.000 \\
solar/10T/S & 0.789 & 25.000 & 0.400 & 26.000 \\
solar/10T/M & 0.787 & 14.000 & 0.297 & 27.000 \\
solar/10T/L & 0.780 & 17.000 & 0.287 & 13.000 \\
solar/H/S & 0.558 & 19.000 & 0.209 & 18.000 \\
solar/H/M & 0.607 & 18.000 & 0.218 & 19.000 \\
solar/H/L & 0.622 & 19.000 & 0.219 & 18.000 \\
solar/D/S & 0.967 & 42.000 & 0.279 & 67.000 \\
solar/W/S & 0.927 & 14.000 & 0.127 & 11.000 \\
hospital/M/S & 0.740 & 10.000 & 0.048 & 8.000 \\
covid\_deaths/D/S & 29.855 & 16.000 & 0.025 & 8.000 \\
us\_births/D/S & 0.305 & 11.000 & 0.015 & 4.000 \\
us\_births/M/S & 0.517 & 18.000 & 0.011 & 15.000 \\
us\_births/W/S & 0.859 & 10.000 & 0.010 & 8.000 \\
saugeen/D/S & 2.746 & 31.000 & 0.336 & 36.000 \\
saugeen/M/S & 0.730 & 57.000 & 0.292 & 54.000 \\
saugeen/W/S & 1.170 & 31.000 & 0.350 & 44.000 \\
temperature\_rain/D/S & 1.305 & 10.000 & 0.537 & 13.000 \\
kdd\_cup\_2018/H/S & 0.796 & 17.000 & 0.320 & 17.000 \\
kdd\_cup\_2018/H/M & 0.899 & 18.000 & 0.364 & 14.000 \\
kdd\_cup\_2018/H/L & 0.873 & 18.000 & 0.381 & 20.000 \\
kdd\_cup\_2018/D/S & 1.200 & 82.000 & 0.378 & 69.000 \\
car\_parts/M/S & 0.844 & 50.000 & 0.956 & 38.000 \\
restaurant/D/S & 0.675 & 12.000 & 0.253 & 13.000 \\
hierarchical\_sales/D/S & 0.740 & 21.000 & 0.570 & 30.000 \\
hierarchical\_sales/W/S & 0.713 & 22.000 & 0.339 & 16.000 \\
loop\_seattle/5T/S & 0.538 & 26.000 & 0.046 & 20.000 \\
loop\_seattle/5T/M & 0.711 & 13.000 & 0.063 & 13.000 \\
loop\_seattle/5T/L & 0.776 & 17.000 & 0.069 & 17.000 \\
loop\_seattle/H/S & 0.755 & 19.000 & 0.052 & 18.000 \\
loop\_seattle/H/M & 0.839 & 21.000 & 0.057 & 18.000 \\
loop\_seattle/H/L & 0.829 & 20.000 & 0.056 & 17.000 \\
loop\_seattle/D/S & 0.858 & 14.000 & 0.041 & 26.000 \\
sz\_taxi/15T/S & 0.540 & 16.000 & 0.199 & 25.000 \\
sz\_taxi/15T/M & 0.532 & 5.000 & 0.200 & 11.000 \\
sz\_taxi/15T/L & 0.503 & 12.000 & 0.195 & 11.000 \\
sz\_taxi/H/S & 0.556 & 8.000 & 0.134 & 21.000 \\
m\_dense/H/S & 0.738 & 12.000 & 0.121 & 10.000 \\
m\_dense/H/M & 0.666 & 12.000 & 0.110 & 8.000 \\
m\_dense/H/L & 0.660 & 9.000 & 0.109 & 8.000 \\
m\_dense/D/S & 0.616 & 17.000 & 0.058 & 15.000 \\
ett1/15T/S & 0.665 & 18.000 & 0.154 & 26.000 \\
ett1/15T/M & 0.958 & 8.000 & 0.223 & 15.000 \\
ett1/15T/L & 0.966 & 9.000 & 0.221 & 11.000 \\
ett1/H/S & 0.791 & 22.000 & 0.171 & 24.000 \\
ett1/H/M & 1.237 & 25.000 & 0.249 & 19.000 \\
ett1/H/L & 1.350 & 35.000 & 0.260 & 24.000 \\
ett1/D/S & 1.626 & 22.000 & 0.263 & 10.000 \\
ett1/W/S & 1.495 & 29.000 & 0.250 & 24.000 \\
ett2/15T/S & 0.715 & 27.000 & 0.062 & 28.000 \\
ett2/15T/M & 0.848 & 24.000 & 0.087 & 13.000 \\
ett2/15T/L & 0.860 & 16.000 & 0.089 & 9.000 \\
ett2/H/S & 0.722 & 30.000 & 0.063 & 26.000 \\
ett2/H/M & 1.006 & 18.000 & 0.101 & 16.000 \\
ett2/H/L & 0.999 & 21.000 & 0.099 & 16.000 \\
ett2/D/S & 1.251 & 11.000 & 0.088 & 10.000 \\
ett2/W/S & 0.861 & 71.000 & 0.086 & 44.000 \\
jena\_weather/10T/S & 0.260 & 24.000 & 0.027 & 25.000 \\
jena\_weather/10T/M & 0.566 & 7.000 & 0.045 & 8.000 \\
jena\_weather/10T/L & 0.614 & 15.000 & 0.046 & 7.000 \\
jena\_weather/H/S & 0.513 & 10.000 & 0.041 & 23.000 \\
jena\_weather/H/M & 0.740 & 13.000 & 0.048 & 10.000 \\
jena\_weather/H/L & 0.862 & 21.000 & 0.051 & 6.000 \\
jena\_weather/D/S & 0.992 & 14.000 & 0.045 & 25.000 \\
bitbrains\_fast\_storage/5T/S & 0.615 & 6.000 & 0.352 & 17.000 \\
bitbrains\_fast\_storage/5T/M & 0.914 & 5.000 & 0.600 & 25.000 \\
bitbrains\_fast\_storage/5T/L & 0.833 & 17.000 & 0.707 & 58.000 \\
bitbrains\_fast\_storage/H/S & 0.965 & 25.000 & 0.615 & 34.000 \\
bitbrains\_rnd/5T/S & 1.576 & 14.000 & 0.374 & 16.000 \\
bitbrains\_rnd/5T/M & 4.327 & 18.000 & 0.625 & 54.000 \\
bitbrains\_rnd/5T/L & 3.250 & 15.000 & 0.592 & 23.000 \\
bitbrains\_rnd/H/S & 5.772 & 21.000 & 0.634 & 60.000 \\
bizitobs\_application/10S/S & 0.935 & 10.000 & 0.010 & 37.000 \\
bizitobs\_application/10S/M & 2.399 & 60.000 & 0.039 & 65.000 \\
bizitobs\_application/10S/L & 3.218 & 57.000 & 0.054 & 67.000 \\
bizitobs\_service/10S/S & 0.680 & 12.000 & 0.010 & 15.000 \\
bizitobs\_service/10S/M & 0.994 & 36.000 & 0.022 & 41.000 \\
bizitobs\_service/10S/L & 1.267 & 13.000 & 0.052 & 50.000 \\
bizitobs\_l2c/5T/S & 0.222 & 21.000 & 0.057 & 18.000 \\
bizitobs\_l2c/5T/M & 0.464 & 21.000 & 0.195 & 20.000 \\
bizitobs\_l2c/5T/L & 0.453 & 14.000 & 0.200 & 10.000 \\
bizitobs\_l2c/H/S & 0.445 & 57.000 & 0.190 & 56.000 \\
bizitobs\_l2c/H/M & 0.476 & 16.000 & 0.228 & 21.000 \\
bizitobs\_l2c/H/L & 0.550 & 29.000 & 0.255 & 31.000 \\
Overall & 0.656 & 19.990 & 0.449 & 22.216 \\
\end{longtable}
\endgroup

\subsection{Context is Key}

Our CiK evaluation uses weighted RCRPS on all 71 tasks, with five
deterministic instances per task (355 evaluations) and 25 forecast
trajectories~\citep{contextiskey}.  Lower is better.
\paragraph{TeeMoE inference.}
The native output path receives up to 168 observed values, their timestamps, the supplied context, and the requested forecast timestamps.
We sample at temperature 1 with thinking disabled. For a horizon of $H$ steps, each trajectory has a budget of $\max(512,64+40H)$ new tokens.
Constrained decoding fixes the timestamp/value syntax; the model predicts the values.
The scorer requires a finite value at every forecast timestamp for each of the 25 trajectories and raises an error for an incomplete or malformed forecast.
Instance seeds are 1--5; separate generation seeds are fixed per task and instance in the reproduction configuration \texttt{configs/native\_cik\_evaluation.json} and its request builder.

Direct prompting (DP) generates forecasts from history and context; direct prompting for forecast correction (CorDP) uses an LLM to adjust a numerical model's forecasts using context~\citep{naiveprompting}.
SW denotes SampleWise.

For CiK placement, we count each model once and retain its best eligible configuration, using the latest reported evaluation of each configuration. For example, the alternative prompting and forecast-correction configurations in \emph{Beyond Na\"ive Prompting}~\citep{naiveprompting} are grouped by model.

\paragraph{Training and evaluation.}
We apply the trained TeeMoE checkpoint directly to each CiK request, without further parameter updates or labelled demonstrations.
The native expert's training sources and example counts are listed in Appendix~\ref{app:native-recipe}; we audited those data for overlap with CiK evaluation instances and found none.
The original CiK comparison likewise includes trained forecasting systems, including UniTime trained on Electricity, which also supplies benchmark tasks~\citep{contextiskey}.
IC-DP includes an additional CiK instance from the same task type in its prompt, supplying that instance's history, context, and ground-truth future as a worked example~\citep[Section~6.2]{naiveprompting}.
TeeMoE receives no such demonstration.
We therefore report IC-DP separately from the zero-shot comparison in Table~\ref{tab:leaderboards}.

\FloatBarrier
\begin{table}[htbp]
\caption{Additional CiK reports and the protocol differences that exclude them
from the full 355-instance comparison.}
\label{tab:cik-omissions}
\centering
\footnotesize
\setlength{\tabcolsep}{4pt}
\begin{tabularx}{\textwidth}{@{}>{\raggedright\arraybackslash}p{0.20\textwidth}>{\raggedright\arraybackslash}p{0.25\textwidth}>{\raggedright\arraybackslash}X@{}}
\toprule
Work or setting & Reported & Reason for exclusion from Table~\ref{tab:leaderboards} \\
\midrule
Beyond Na\"ive Prompting, IC-DP
  & Gemini-2.5-Pro $0.100\!\pm\!0.001$; Llama-405B $0.129\!\pm\!0.004$
  & One-shot evaluation: IC-DP supplies a labelled instance from the same CiK
    task type, including its ground-truth future. We include DP/CorDP without
    demonstrations in Table~\ref{tab:leaderboards} and report IC-DP separately~\citep{naiveprompting}. \\
TimeClaw
  & Average RCRPS $0.115$
  & This score covers the held-out portion of a benchmark train/test split,
    rather than the full 355-instance panel ranked in Table~\ref{tab:leaderboards}.
    Scores over these different evaluation populations are not directly
    comparable~\citep{timeclaw}. \\
TimeClaw comparison rows
  & Multi-Agent Reflection $0.129$; TS-Agent $0.142$; TSci $0.145$
  & These values are reported in the same TimeClaw table and inherit its
    held-out evaluation subset~\citep{timeclaw}. \\
Multimodal forecasting study
  & MAE, MSE, WQL, and ordinary CRPS
  & The arXiv v2 CiK experiments explicitly omit RCRPS, so none of these metrics is the
    canonical ranking statistic~\citep{multimodalityforecasting}. \\
TsLLM
  & sMAPE $64.500\%$; MASE $64.700\%$
  & No RCRPS is reported and the evaluated instances are not specified; the point
    metrics cannot be converted into weighted RCRPS~\citep{tsllm}. \\
CodeCast
  & MSE $22{,}038.000$; MAE $30.520$, plus ELO/rank/win rate
  & These are point-forecast or preference metrics rather than probabilistic
    RCRPS~\citep{codecast}. \\
Dr-CiK
  & sCRPS on 240 Dr-CiK tasks
  & Dr-CiK is a distinct benchmark.  Although 199 generated tasks draw on CiK
    resources, its task set and metric are not the canonical CiK
    protocol~\citep{drcik}. \\
\bottomrule
\end{tabularx}
\end{table}
\FloatBarrier

\paragraph{Repeated Llama-3.1-405B reports.}
The original CiK paper reported $0.159\!\pm\!0.008$; the maintainers' July 2025
table reported $0.143\!\pm\!0.006$ after a covariate-scaling correction\footnote{\url{https://github.com/ServiceNow/context-is-key-forecasting/blob/main/README.md}}; and a
later evaluation reported $0.173\!\pm\!0.003$
\citep{contextiskey,naiveprompting}. The main panel uses the latest reported
evaluation of this direct-prompting configuration, $0.173$.

\subsection{TimeSeriesExam}

We use the official v1.1 release of TimeSeriesExam, containing 746 questions: pattern recognition 362, noise understanding 84, anomaly detection 108, similarity analysis 120, and causality analysis 72~\citep{timeseriesexam,tsev11}.
The 746 records contain 741 distinct full prompts.
This revision corrects ambiguous or incorrect question cases and standardizes formatting.
Our model uses the analysis input format in Appendix~\ref{app:analysis-prompts} and generates one answer per question.
Decoding is greedy, with thinking disabled, generation seed 1, and a 64-token output limit. Responses are generated without constrained decoding and scored as returned; empty or unmatched answers are incorrect, and length-capped responses remain in the denominator. Exact final-line labelled-option matching and flexible scoring agree on all 746 composed outputs.

The version-matched comparison includes the official dataset-card baselines and the direct, code-only, and hybrid configurations reported by \citet{codingagents}.
Their hybrid setup combines raw values with iterative Python access; its GPT-oss-120B result is 78.000\%, compared with our 78.552\%.
The table preserves each published configuration's model size, input representation, and tool access.

\FloatBarrier
\begin{table}[htbp]
\caption{Other TSE reports and their relationship to the v1.1 comparison.
Reported values are not converted into full-population accuracy when counts or averaging rules are unavailable.}
\label{tab:tse-omissions}
\centering
\footnotesize
\setlength{\tabcolsep}{4pt}
\begin{tabularx}{\textwidth}{@{}>{\raggedright\arraybackslash}p{0.20\textwidth}>{\raggedright\arraybackslash}p{0.25\textwidth}>{\raggedright\arraybackslash}X@{}}
\toprule
Work or setting & Reported & Comparison scope\\
\midrule
Original TSE v1.0 & Category accuracies, including strong one-shot image/text baselines & The original 763-question population differs from v1.1; its rounded category results are not inserted into the v1.1 table~\citep{timeseriesexam}.\\
TSEA & GPT-4o 73.000\%; Gemini-2.5-Pro 71.000\% & Category counts describe 763 questions, not 746~\citep{tseagent}.\\
TS-Reasoner report & GPT-4.1 vision 67.890\%; ChatTS-14B 56.360\%; TS-Reasoner-7B 54.830\% & The category denominators are consistent with the older 763-question population, without establishing question-level identity~\citep{tsreasoner}.\\
TsLLM & Easy 87.300\%; hard 85.500\% & The inspected arXiv v2 does not specify the release, partition sizes, or overall accuracy, preventing placement on the full v1.1 ranking~\citep{tsllm}.\\
ITFormer & High category accuracies & TSE fine-tuning is reported without an established held-out split, release, or overall accuracy~\citep{itformer}.\\
TimeOmni-1 & 47.800\% average & Uses 746 questions, but reports an equal-category mean rather than all-question accuracy~\citep{timeomni}.\\
VeriTime & Qwen3 47.270\% & Release and item count are unspecified~\citep{veritime}.\\
Domain-oriented TS-Reasoner & Category radar & No exact overall score or confirmed release is provided~\citep{domainreasoner}.\\
\bottomrule
\end{tabularx}
\end{table}
\FloatBarrier

\subsection{Model references for Table~\ref{tab:leaderboards}}
\label{app:table1-model-references}

\paragraph{Forecasting systems.}
For STRIDE + Synapse, we cite the STRIDE~\citep{stride} and Synapse~\citep{synapse} papers.
EXAONE Forecast Agent, TW3Cast, and LS-MoE are documented in the official GIFT-Eval submission records~\citep{gifteval_results}.

\paragraph{Numerical foundation models.}
The selected models are TimesFM-3~\citep{timesfm3}, Chronos-2~\citep{chronos2}, and Toto-2.0-2.5B~\citep{toto2}.
The CorDP configurations additionally use Chronos-Large~\citep{chronos}.

\paragraph{Time-series language models.}
We cite TimeOmni-VL~\citep{timeomnivl}, TS-Reasoner-7B~\citep{tsreasoner}, ChatTS-14B~\citep{chatts}, and OpenTSLM-SoftPrompt with Llama 3.2 1B~\citep{opentslm}.

\paragraph{General-purpose language models.}
The CiK comparisons use Gemini-2.5-Pro~\citep{gemini25}, Claude-Sonnet-4.5~\citep{claude45}, GPT-5.2~\citep{gpt52}, and Llama-3.1-405B-Instruct~\citep{llama3}, with the DP and CorDP configurations of \citet{naiveprompting}.
The TSE comparisons use GPT-oss-120B~\citep{gptoss} in code-only and hybrid configurations, GPT-4o~\citep{gpt4o} with image input, and Qwen3-Next-80B~\citep{qwen3next} in a hybrid configuration; evaluation configurations follow the official TSE dataset card~\citep{tsev11} and \citet{codingagents}.

\clearpage
\section{System and user prompts}
\label{app:shared-prompt}

The following instruction is shared across textual requests.
Only the example-specific user payload changes.

\subsection{Shared system instruction}

The instruction below is reproduced verbatim.
The controller selects the output path from its aggregation weight, learned with numerical-versus-text format supervision (Appendix~\ref{app:output-selection}).
The instruction calls the reference forecast a ``protected aggregate.''

\begin{quote}\small
First identify whether the raw request is time-series analysis or forecasting. If it is forecasting, choose structured aggregation versus native contextual forecasting from the supplied evidence, never from task identity or the mere presence of candidates.

TIME-SERIES ANALYSIS: Derive the answer from the raw series and relevant supplied measurements. Use only diagnostics that bear on the requested property; do not let irrelevant scale, offset, amplitude, visual complexity, terminology, or incidental attributes override the numerical evidence. When options are present, decide which option text the evidence supports before mapping it to its label.

STRUCTURED AGGREGATION: Compare supplied candidate forecast distributions and any protected aggregate. Make only calibrated, numerically supported changes through the structured forecasting output.

NATIVE CONTEXTUAL FORECASTING: Generate one coherent plausible sample from the conditional future. Start from a continuation that preserves the target history's level, dynamics, dependence, seasonality, and uncertainty, and change it only where supplied context provides supported future evidence. Enforce exact values, equations, and bounds, apply interventions with their supported timing and magnitude, and ignore context that is merely descriptive or irrelevant.

Return only the requested output. For a multiple-choice analysis, return exactly one line in the form \texttt{LETTER) exact option text}; do not explain or repeat the options. For a native forecast, return only (timestamp, value) pairs inside \texttt{<forecast>} and \texttt{</forecast>}.
\end{quote}

\subsection{Analysis inputs}
\label{app:analysis-prompts}

Following TS-Agent, TS-Reasoner, and coding agents in supplying numerical information for time-series reasoning~\citep{tsagent,tsreasoner,codingagents}, our analysis requests include raw observations and fixed numerical summaries of their level and variability, local and overall trends, temporal dependence and periodicity, and relationships between series.
The standalone analysis expert and the composed model use this input format in both training and evaluation.

The summaries are computed solely from the observed time-series values, independently of the question, answer options, and reference answer.
TS-Agent provides summary-statistics and signal-analysis tools~\citep{tsagent}.
The GPT-oss-120B hybrid coding agent combines raw observations with Python computations~\citep{codingagents}.
TeeMoE supplies a fixed set of computations in the prompt; these agents select computations during inference.
Numerical input metadata also appears in TSLMs: ChatTS and TS-Reasoner supply offset and scale values, and OpenTSLM supplies mean, standard deviation, and temporal information~\citep{chatts,tsreasoner,opentslm}.
The unadapted-backbone control uses the same inputs when assessing adaptation (Appendix~\ref{app:raw-backbone}).

Optional source-supplied definitions and clarifications accompany the question.
Raw observations are rendered with four significant digits. The reproduction code implements this input preparation and identifies the selected training examples.

\subsection{User-message templates}

The four templates below render examples with separate history, context,
question, and answer-format fields as applicable. They accompany the same
system instruction in training and evaluation.
Braced fields are filled with the example's supplied data;
they do not identify a benchmark or select an expert. Forecasting requests
share the same history and future-timestamp format, with an added context block
when context is available. Analysis requests share the same evidence and
question fields, with answer options when provided.
Appendix~\ref{app:numerical-interface} describes the aggregation expert's numerical interface.

\noindent\begin{minipage}{\linewidth}
\paragraph{Forecasting without context.}
{\small
\begin{verbatim}
I have a time series forecasting task for you.

Historical time series in (timestamp, value) format:
<history>
{history_evidence}
</history>

Future timestamps:
{prediction_points}
\end{verbatim}
}

\end{minipage}
\par\medskip
\noindent\begin{minipage}{\linewidth}
\paragraph{Forecasting with context.}
{\small
\begin{verbatim}
I have a time series forecasting task for you.

Context:
<context>
{context}
</context>

Historical time series in (timestamp, value) format:
<history>
{history_evidence}
</history>

Future timestamps:
{prediction_points}
\end{verbatim}
}

\end{minipage}
\par\medskip
\noindent\begin{minipage}{\linewidth}
\paragraph{Multiple-choice analysis.}
{\small
\begin{verbatim}
Analyze the supplied time-series evidence.

Evidence:
{evidence}

Question:
{question}

Options:
{options}

Requested output:
{output_schema}
\end{verbatim}
}

\end{minipage}
\par\medskip
\noindent\begin{minipage}{\linewidth}
\paragraph{Free-form analysis.}
{\small
\begin{verbatim}
Analyze the supplied time-series evidence.

Evidence:
{evidence}

Question:
{question}

Requested output:
{output_schema}
\end{verbatim}
}

\end{minipage}
\par\medskip

For multiple-choice analysis, the default output schema is
\texttt{Return exactly one line as LETTER) exact option text.}
For free-form analysis, it is
\texttt{Answer the analytical question directly.}
When the source example specifies another answer format, its requested format
fills this field.

\subsection{Example of routing and expert inputs}
\label{app:routing-example}

\begin{figure}[htbp]
\centering
\includegraphics[width=\linewidth]{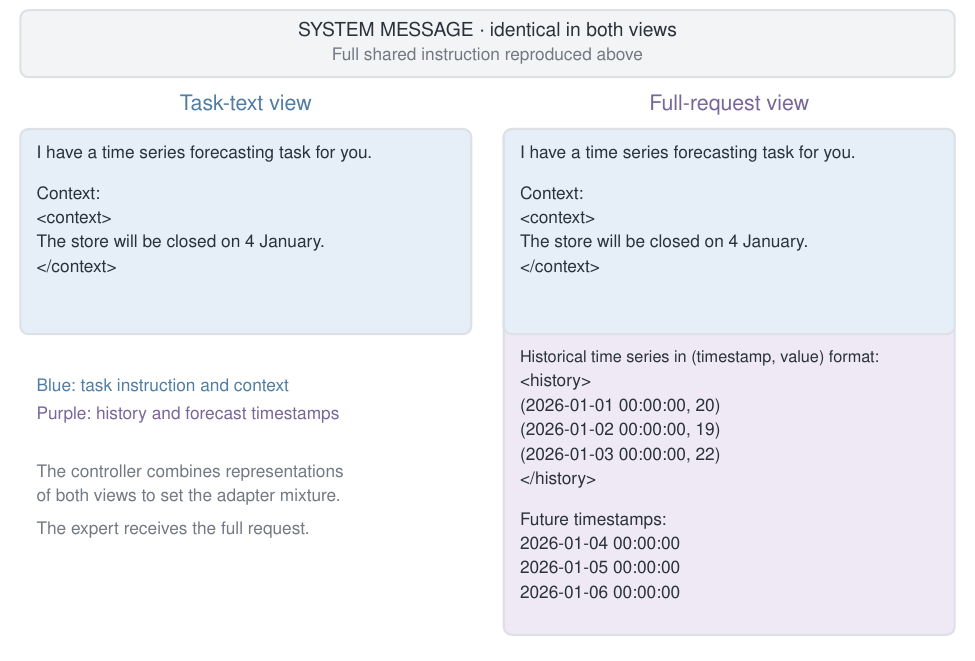}
\caption{Task-text and full-request views of an illustrative contextual
forecasting request. Blue fields appear in both views; the full view also
contains the purple history and forecast-timestamp fields. The controller
combines both representations, and the expert receives the full request.
The shared system instruction is unchanged.}
\label{fig:routing-example}
\end{figure}

Figure~\ref{fig:routing-example} shows the task-text and full-request views of a
contextual forecasting request. The adapter-disabled backbone encodes both
views, whose representations are combined for the controller as defined in
Equation~\ref{eq:controller-input}. The subsequent expert pass receives the full
request and applies the resulting adapter mixture. All views retain the shared
system instruction. The illustrated text-output path generates the
forecast directly. If numerical output is selected, foundation-model forecasts
are obtained from the observed history and passed through the numerical
connector described in Appendix~\ref{app:numerical-interface}.

\FloatBarrier

\section{Evaluation of comparison models}
\label{app:baseline-evaluation}

Asterisks in Table~\ref{tab:leaderboards} mark our evaluations of released
comparison models. We retain their model-specific interfaces; the following
protocols specify the inputs, generation settings, and scoring used for each.
We evaluate the released checkpoints on the benchmark populations specified
above; author-reported results are identified separately in
Appendix~\ref{app:leaderboard-audit}.

\subsection{GIFT-Eval compute budget}
\label{app:gift-runtime}

We mark a GIFT comparison OOT (out of time) when its projected full evaluation
exceeds our predeclared budget of 1,000 allocated H100 GPU-hours per model.
We estimate the cost of generating 25 trajectories for each of the 371,330
evaluation windows using eight requests sampled without access to future
values: two from each of the four nonempty horizon strata (1--12, 13--48,
49--96, and 385+ steps).
For each stratum, we multiply the mean measured request cost by its population
and sum across strata. A faster-request sensitivity substitutes the less
expensive of the two requests in every stratum; OOT requires this estimate
also to exceed the budget.

The comparison text models retain the forecasting interfaces described below, with one
draw per batch to accommodate long GIFT inputs. We measure two completed draws
per request and extrapolate to 25; TimeOmni-VL measures all 25 draws through
its image-generation interface. OpenTSLM receives
the latest 4,096 history points, matching its encoder capacity. Our standalone
native forecasting and analysis experts use their canonical forecasting prompt
and latest 168 history points, measuring two concurrent draws per request in
vLLM and extrapolating their amortized cost to 25 draws.
Table~\ref{tab:gift-runtime} reports rounded, sample-based projections under these execution settings, excluding model loading.

\begin{table}[htbp]
\centering
\caption{Projected GIFT-Eval cost in allocated H100 GPU-hours for 25 trajectories
per window, rounded to two significant figures. All six profiled models
exceed the 1,000-hour budget under both estimates.}
\label{tab:gift-runtime}
\begin{tabular}{lrr}
\toprule
Model & Mean-based projection & Faster-request sensitivity \\
\midrule
OpenTSLM SP Llama 3.2 1B & 35,000 & 28,000 \\
ChatTS-14B & 450,000 & 330,000 \\
TS-Reasoner-7B & 500,000 & 350,000 \\
TimeOmni-VL & 100,000 & 92,000 \\
\midrule
Native forecasting expert & 130,000 & 110,000 \\
Analysis expert & 140,000 & 110,000 \\
\bottomrule
\end{tabular}
\end{table}

\subsection{Context is Key}

All local CiK evaluations cover the same 355 instances and use 25 trajectories
per instance with the canonical scorer.
Following CiK's official result-compilation script, the table reports RCRPS with a per-instance cap of 5; uncapped values below
describe sensitivity to unusually large errors.
Table~\ref{tab:cik-cap-sensitivity} lists both aggregates for every local CiK contender and ablation reported in the paper, alongside the published comparison scores.
The two columns score the same predictions; no forecasts are regenerated or discarded.
TeeMoE's main score is identical with and without capping.
Model-specific inputs and trajectory construction are detailed next.

\begin{table}[tp]
\centering
\caption{CiK RCRPS with and without the official per-instance cap of 5 (lower is better). Local results use all 355 instances and the same saved 25 trajectories for both columns. Published Table~\ref{tab:leaderboards} contenders are listed separately: their reported scores are retained, but paired capped/uncapped values are unavailable. N/A denotes an unsupported full-benchmark evaluation.}
\label{tab:cik-cap-sensitivity}
\footnotesize
\begin{tabular}{@{}lrr@{}}
\toprule
Configuration & Capped & Uncapped \\
\midrule
\multicolumn{3}{@{}l}{\textit{TeeMoE and adapter controls}} \\
OpenTSLM TeeMoE & 0.115 & 0.115 \\
Qwen3.6-27B (no adapters) & 0.151 & 0.169 \\
Aggregation expert (text transfer) & 0.154 & 0.168 \\
Analysis expert & 0.138 & 0.138 \\
Native forecasting expert & 0.123 & 0.123 \\
TeeMoE (top-1 routing) & 0.123 & 0.123 \\
Equal-weight adapters & 0.119 & 0.119 \\
Full-strength adapters & 0.121 & 0.121 \\
Joint training & 0.123 & 0.123 \\
\midrule
\multicolumn{3}{@{}l}{\textit{Numerical foundation models}} \\
TimesFM-3 & 0.491 & 0.530 \\
Chronos-2 & 0.358 & 0.358 \\
Toto-2.0-2.5B & 0.333 & 0.333 \\
\midrule
\multicolumn{3}{@{}l}{\textit{Time-series language models}} \\
TimeOmni-VL & 0.386 & 0.386 \\
TS-Reasoner-7B & 0.374 & 0.374 \\
ChatTS-14B & 0.322 & 15.429 \\
OpenTSLM SP Llama 3.2 1B & 0.779 & 2.210 \\
\midrule
\multicolumn{3}{@{}l}{\textit{Numerical ensembles}} \\
Equal-weight FM ensemble (13) & 0.283 & 0.283 \\
Equal-weight FM ensemble (8) & 0.283 & 0.283 \\
XGBoost-weighted FM ensemble (8) & 0.286 & 0.286 \\
XGBoost + Toto-FnF (reference blend) & 0.292 & 0.292 \\
Toto-FnF alone & N/A & N/A \\
\midrule
\multicolumn{3}{@{}l}{\textit{Published contenders (reported RCRPS; paired breakdown unavailable)}} \\
DP: Gemini-2.5-Pro & \multicolumn{2}{c}{0.108 (reported)} \\
SW-CorDP: Claude-Sonnet-4.5 + Chronos-Large & \multicolumn{2}{c}{0.110 (reported)} \\
Median-CorDP: GPT-5.2 + Chronos-Large & \multicolumn{2}{c}{0.167 (reported)} \\
DP: Llama-3.1-405B-Instruct & \multicolumn{2}{c}{0.173 (reported)} \\
\bottomrule
\end{tabular}
\end{table}

\paragraph{Numerical foundation models on CiK.}
Chronos-2, TimesFM-3, and Toto-2.0-2.5B receive the observed target history and forecast horizon, without textual context; their released predictors do not accept natural-language instructions.
These released prediction interfaces provide marginal quantiles, which we convert to the sample trajectories required by CiK.
We use each model's native quantile grid and construct 25 trajectories by independently sampling the linearly interpolated inverse CDF at each future step, with constant tails beyond the outermost quantiles.
Crossed quantiles are sorted before sampling.

\paragraph{Local TimeOmni-VL forecasting evaluation.}
We evaluate the released TimeOmni-VL checkpoint on the same 355 instances through
the authors' official forecasting interface: history-image conversion, reasoning
and image generation, and forecast reconstruction. A local prompt supplies the observed history, full textual
context and forecast timestamps. We retain 25 native image-derived trajectories
per instance, using 50 image-generation steps and text/image guidance weights
of 4/2. This local CiK transfer scores 0.386 under both capped and uncapped
RCRPS. Generation uses 99.96 allocated H100 GPU-hours.

\paragraph{Local ChatTS forecasting evaluation.}
We evaluate ChatTS-14B on all 355 instances through its released signal
processor and chat interface, providing the full context and actual forecast
timestamps. We draw 25 trajectories per instance in batches of five at
temperature 0.7, constraining timestamp/value syntax without imposing
numerical bounds. The transfer achieves an RCRPS of 0.322 with a per-instance
RCRPS cap of 5; the uncapped aggregate is 15.429, with three pressure-task
outliers. The 25 generated trajectories are identical for 140 instances.

\paragraph{Local TS-Reasoner forecasting evaluation.}
We evaluate TS-Reasoner-7B on the same 355 instances through its released
TimesFM/SP signal interface and chat template, providing full context and
actual forecast timestamps. We draw 25 trajectories in batches of five at
temperature 0.7 with syntax-only timestamp/value constraints. The transfer scores 0.374 under
both capped and uncapped RCRPS. The 25 trajectories are identical for 16 instances.

\paragraph{Local OpenTSLM forecasting evaluation.}
We use OpenTSLM's final ECG SoftPrompt checkpoint for both CiK and TSE: the authors' released inference checkpoint after the complete five-stage curriculum, incorporating all preceding training stages~\citep{opentslm}.
For CiK, we use the authors' signal normalization and provide textual context and
forecast timestamps. We draw 25 trajectories in batches of five at temperature
0.6 and top-$p$ 0.9, with a local numerical-output prompt and syntax-only JSON
constraints. This forecasting transfer scores 0.779 capped RCRPS and
2.210 uncapped; six instances exceed the cap of 5. There are 7--24 distinct
trajectories per instance. Generation uses
2.780 H100 GPU-hours.

\FloatBarrier
\subsection{TimeSeriesExam}

TimeSeriesExam provides two official scoring modes. Flexible scoring checks
for the correct option letter followed by its answer text anywhere in the
response; strict scoring checks for the answer text in the final line. Both
ignore case, and their different matching rules mean either can give a higher
score. We evaluate all competitor models with both official TSE scorers and
report the higher accuracy, using the same 746 saved responses per model.
\method{} and its ablations retain the default flexible scorer.
These evaluations retain model-specific inputs and generation settings.

\paragraph{Local TS-Reasoner evaluation.}
We evaluate the released TS-Reasoner-7B checkpoint on all 746 v1.1 questions
using the prompt published in its Appendix E, Figure 13~\citep{tsreasoner},
with its native TimesFM-based encoder and chat interface.
Greedy decoding with a 512-token limit yields 368/746 (49.330\%) under
official flexible scoring and 393/746 (52.681\%) under strict scoring; the table
reports the latter. No response reaches the length limit.
The authors' transformed evaluation inputs are unavailable, so we reconstruct
their published prompt for v1.1.

\paragraph{Local ChatTS evaluation.}
ChatTS-14B receives the benchmark's question wrapper, supplied definitions and
hints through its native chat interface, with the numerical signals processed
by its released encoder.
We sample one response at temperature 0.2 with a 1,024-token limit. Official
flexible scoring gives 376/746 (50.402\%), while strict scoring gives the
reported 429/746 (57.507\%).
The single length-capped response remains in the denominator.

\paragraph{Local OpenTSLM evaluation.}
We evaluate OpenTSLM SP Llama 3.2 1B using the same final ECG SoftPrompt checkpoint,
including the encoder, projector, and LoRA weights, on all 746 questions.
The ECG stage completes the authors' five-stage curriculum; this checkpoint is intended for inference after the full training sequence~\citep{opentslm}.
Its numerical TSQA-style interface receives normalized signals, questions,
options, hints, and concept definitions. This local prompt
protocol is distinct from the official TSE wrapper and the authors' ECG task.
With one sampling seed, temperature 0.6, top-$p$ 0.9, and a 500-token limit,
official flexible scoring gives 113/746 (15.147\%). All 28 truncated responses
remain included. Official strict matching gives the reported 200/746
(26.810\%).

\paragraph{Local TimeOmni-VL evaluation.}
\label{app:timeomni-tse}
We use TimeOmni-VL's official text-only reasoning interface and thinking system
prompt for all 746 questions, paired with the benchmark
question wrapper, with greedy decoding and a 1,024-token limit.
The authors also evaluate text-only time-series reasoning in their paper~\citep[Appendix E.3]{timeomnivl}.
Official flexible scoring yields the reported 110/746 (14.745\%); strict scoring
gives 35/746 (4.692\%). All 31 length-capped responses remain included.
Inspection of the saved responses reveals bare-letter answers without the
requested option text, unfinished reasoning, and occasional contradictions
between the explanation and final choice. Responses can also quote several
options during reasoning, allowing the flexible scorer to match an option
other than the final choice. We retain all responses and report the higher
of the two official scores, following the same rule used for every local competitor.

\section{Reproducibility and responsible use}
\label{app:responsible-use}

\subsection{Release and source assets}
\label{app:release}

The codebase is available on \href{https://github.com/OpenTSLM/OpenTSLM-TeeMoE}{GitHub}, and model checkpoints are available on \href{https://huggingface.co/OpenTSLM/TeeMoE}{Hugging Face}.
It contains the implementation, selected configurations, and guides for source preparation, expert training, composition, ablations, and comparison-model evaluation. Fixed source coordinates and membership metadata specify the training selections. Datasets, forecast caches, and learned weights are downloaded or generated through these workflows and stored separately from the source checkout.

\subsection{Computational resources}
\label{app:compute}

Experiments used a cloud-hosted Linux server with eight NVIDIA H100 GPUs with 80\,GB of memory each, 128 logical Intel Xeon Platinum 8468 CPUs, approximately 1.5\,TiB host RAM, and 3.7\,TiB local storage.
Table~\ref{tab:compute} reports cache-preparation and training costs in allocated H100 GPU-hours.

\begin{table}[htbp]
\centering
\caption{Cache preparation and training costs in H100 GPU-hours.}
\label{tab:compute}
\begin{tabular}{@{}lr@{}}
\toprule
Component & H100 GPU-hours\\
\midrule
FM forecast caches, training & 58.530\\
FM forecast caches, evaluation & 16.770\\
\midrule
XGBoost, full-data fit and ten cross-fitting folds & 1.390\\
Aggregation editor, 4,096 training examples & 0.480\\
Native forecasting, 20,000 examples & 21.839\\
Analysis, 12,000 examples & 6.930\\
Composition controller, 1,000 examples & 1.508\\
\bottomrule
\end{tabular}
\end{table}

The controller entry measures its training loop, excluding model loading and validation.
The XGBoost entry sums the selected full-data and ten cross-fitting fits; it excludes population loading, forecast caching, cross-fitted ensemble prediction, scalar fitting, and preparation of the editor's training inputs.

With foundation-model forecasts and request-routing states already cached, full GIFT-Eval evaluation used 8.000 H100 GPU-hours for the composed model; cache construction is additional.
The composed model's full CiK evaluation used 5.780 H100 GPU-hours, excluding CPU-only scoring.
TimeSeriesExam model loading and generation used 0.550 H100 GPU-hours.
We estimate 130 H100 GPU-hours for cache preparation, training, and evaluation of the composed model, and 300 H100 GPU-hours including the original local baselines and ablations; these estimates exclude the subsequent joint-training baseline and additional development runs.
Data downloads, CPU preprocessing, and CPU scoring use the host resources described above.

\subsection{Modular and joint training costs}
\label{app:modular-compute}

We account separately for shared numerical preparation, model training, and capability-specific revision.
The expert and controller training costs in Table~\ref{tab:compute} sum to
\[
 C_{\mathrm{modular}}\approx0.480+21.839+6.930+1.508\approx30.756.
\]
The completed joint run uses eight H100s for 26,395.480 seconds from the start of its training loop through final checkpointing, giving
\[
 C_{\mathrm{joint}}=8\times26{,}395.480/3600=58.657.
\]
Its separate output selector fits on cached request representations on the CPU in approximately 29 seconds; this adds no GPU training time.
These loop timings exclude model loading and request-feature extraction.

Both recipes use the same training forecast caches and fitted ensemble, costing
\[
 C_{\mathrm{prepare}}=58.530+1.390=59.920.
\]
Table~\ref{tab:training-revision-cost} excludes this shared preparation and the shared evaluation forecast caches; their costs are reported separately in Table~\ref{tab:compute}.

Joint training processes 4,512 updates on the same 4,096 aggregation, 20,000 native, and 12,000 analysis example presentations.
Each update evaluates task-specific losses and synchronizes gradients across GPUs whose examples can differ in length and output mechanism.
Equal presentation counts therefore do not imply equal GPU-hours: synchronous execution waits for the longest rank-local workload and combines distinct loss computations within an update.
The measured training-loop ratio is 1.907 (Appendix~\ref{app:joint-comparison-scope} gives the joint recipe).

\subsection{Potential impacts}
\label{app:impacts}

A shared forecasting and analysis model could support applications such as demand planning and infrastructure monitoring.
Incorrect forecasts or interpretations may nevertheless lead to harmful decisions, particularly in consequential settings, while sensitive time-series inputs raise privacy concerns.
Deployment therefore requires validation on the intended population, appropriate data protections, and human oversight.
Benchmark performance alone does not establish fitness for autonomous use in high-stakes applications.


\begin{thebibliography}{78}
\providecommand{\natexlab}[1]{#1}
\providecommand{\url}[1]{\texttt{#1}}
\expandafter\ifx\csname urlstyle\endcsname\relax
  \providecommand{\doi}[1]{doi: #1}\else
  \providecommand{\doi}{doi: \begingroup \urlstyle{rm}\Url}\fi

\bibitem[Ahamed et~al.(2026)Ahamed, Parmar, Goyal, Li, Cheng, Pfister, and
  Yoon]{stride}
Md~Atik Ahamed, Mihir Parmar, Palash Goyal, Chun-Liang Li, Qiang Cheng, Tomas
  Pfister, and Jinsung Yoon.
\newblock Reasoning-aware training for time series forecasting.
\newblock \emph{arXiv preprint arXiv:2605.08625}, 2026.
\newblock \doi{10.48550/arXiv.2605.08625}.

\bibitem[Aksu et~al.(2024)Aksu, Woo, Liu, Liu, Liu, Savarese, Xiong, and
  Sahoo]{gifteval}
Taha Aksu, Gerald Woo, Juncheng Liu, Xu~Liu, Chenghao Liu, Silvio Savarese,
  Caiming Xiong, and Doyen Sahoo.
\newblock {GIFT-Eval}: A benchmark for general time series forecasting model
  evaluation.
\newblock \emph{arXiv preprint arXiv:2410.10393}, 2024.
\newblock URL \url{https://arxiv.org/abs/2410.10393v2}.

\bibitem[Ansari et~al.(2024)Ansari, Stella, Turkmen, Zhang, Mercado, Shen,
  Shchur, Rangapuram, Pineda~Arango, Kapoor, Zschiegner, Maddix, Wang, Mahoney,
  Torkkola, Wilson, Bohlke-Schneider, and Wang]{chronos}
Abdul~Fatir Ansari, Lorenzo Stella, Caner Turkmen, Xiyuan Zhang, Pedro Mercado,
  Huibin Shen, Oleksandr Shchur, Syama~Sundar Rangapuram, Sebastian
  Pineda~Arango, Shubham Kapoor, Jasper Zschiegner, Danielle~C. Maddix, Hao
  Wang, Michael~W. Mahoney, Kari Torkkola, Andrew~Gordon Wilson, Michael
  Bohlke-Schneider, and Yuyang Wang.
\newblock Chronos: Learning the language of time series.
\newblock \emph{Transactions on Machine Learning Research}, 2024.
\newblock ISSN 2835-8856.
\newblock URL \url{https://openreview.net/forum?id=gerNCVqqtR}.

\bibitem[Ansari et~al.(2025)Ansari, Shchur, K{\"u}ken, Auer, Han, Mercado,
  Rangapuram, Shen, Stella, Zhang, Goswami, Kapoor, Maddix, Guerron, Hu, Yin,
  Erickson, Desai, Wang, Rangwala, Karypis, Wang, and
  Bohlke-Schneider]{chronos2}
Abdul~Fatir Ansari, Oleksandr Shchur, Jaris K{\"u}ken, Andreas Auer, Boran Han,
  Pedro Mercado, Syama~Sundar Rangapuram, Huibin Shen, Lorenzo Stella, Xiyuan
  Zhang, Mononito Goswami, Shubham Kapoor, Danielle~C. Maddix, Pablo Guerron,
  Tony Hu, Junming Yin, Nick Erickson, Prateek~Mutalik Desai, Hao Wang, Huzefa
  Rangwala, George Karypis, Yuyang Wang, and Michael Bohlke-Schneider.
\newblock {Chronos-2}: From univariate to universal forecasting.
\newblock \emph{arXiv preprint arXiv:2510.15821}, 2025.
\newblock URL \url{https://arxiv.org/abs/2510.15821}.

\bibitem[{Anthropic}(2025)]{claude45}
{Anthropic}.
\newblock {Claude Sonnet 4.5 System Card}.
\newblock System card, 2025.
\newblock URL
  \url{https://assets.anthropic.com/m/12f214efcc2f457a/original/Claude-Sonnet-4-5-System-Card.pdf}.

\bibitem[Ashok et~al.(2026)Ashok, Williams, Zheng, Rish, Chapados, Marcotte,
  Zantedeschi, and Drouin]{naiveprompting}
Arjun Ashok, Andrew~Robert Williams, Vincent~Zhihao Zheng, Irina Rish, Nicolas
  Chapados, {\'E}tienne Marcotte, Valentina Zantedeschi, and Alexandre Drouin.
\newblock Beyond na{\"i}ve prompting: Strategies for improved context-aided
  forecasting with {LLMs}.
\newblock \emph{Transactions on Machine Learning Research}, 2026.
\newblock URL \url{https://arxiv.org/abs/2508.09904}.

\bibitem[Auer et~al.(2025)Auer, Podest, Klotz, B{\"o}ck, Klambauer, and
  Hochreiter]{tirex}
Andreas Auer, Patrick Podest, Daniel Klotz, Sebastian B{\"o}ck, G{\"u}nter
  Klambauer, and Sepp Hochreiter.
\newblock {TiRex}: Zero-shot forecasting across long and short horizons with
  enhanced in-context learning.
\newblock In \emph{Advances in Neural Information Processing Systems}, 2025.
\newblock URL \url{https://arxiv.org/abs/2505.23719}.

\bibitem[{Auton Lab}(2025)]{tsev11}
{Auton Lab}.
\newblock {TimeSeriesExam-1} dataset card.
\newblock Hugging Face, 2025.
\newblock URL \url{https://huggingface.co/datasets/AutonLab/TimeSeriesExam1}.
\newblock v1.1 released March 12, 2025.

\bibitem[Buehler \& Buehler(2024)Buehler and Buehler]{xlora}
Eric~L. Buehler and Markus~J. Buehler.
\newblock {X-LoRA}: Mixture of low-rank adapter experts, a flexible framework
  for large language models with applications in protein mechanics and
  molecular design.
\newblock \emph{APL Machine Learning}, 2\penalty0 (2):\penalty0 026119, 2024.
\newblock \doi{10.1063/5.0203126}.
\newblock URL \url{https://doi.org/10.1063/5.0203126}.

\bibitem[Cai et~al.(2024)Cai, Choudhry, Goswami, and Dubrawski]{timeseriesexam}
Yifu Cai, Arjun Choudhry, Mononito Goswami, and Artur Dubrawski.
\newblock {TimeSeriesExam}: A time series understanding exam.
\newblock In \emph{NeurIPS Workshop on Time Series in the Age of Large Models},
  2024.
\newblock URL \url{https://arxiv.org/abs/2410.14752}.

\bibitem[Cao et~al.(2024)Cao, Jia, Arik, Pfister, Zheng, Ye, and Liu]{tempo}
Defu Cao, Furong Jia, Sercan Arik, Tomas Pfister, Yixiang Zheng, Wen Ye, and
  Yan Liu.
\newblock {TEMPO}: Prompt-based generative pre-trained transformer for time
  series forecasting.
\newblock In \emph{International Conference on Learning Representations}, pp.\
  18546--18578, 2024.
\newblock URL
  \url{https://proceedings.iclr.cc/paper_files/paper/2024/file/5132940b1bced8a7b28e9695d49d435a-Paper-Conference.pdf}.

\bibitem[{CastStar}(2026)]{caststar}
{CastStar}.
\newblock {CastStar}.
\newblock Hugging Face model card and reproduction artifacts, 2026.
\newblock URL \url{https://huggingface.co/CastStar/CastStar}.

\bibitem[Chen \& Guestrin(2016)Chen and Guestrin]{xgboost}
Tianqi Chen and Carlos Guestrin.
\newblock {XGBoost}: A scalable tree boosting system.
\newblock In \emph{Proceedings of the 22nd ACM SIGKDD International Conference
  on Knowledge Discovery and Data Mining}, KDD '16, pp.\  785--794. ACM, August
  2016.
\newblock \doi{10.1145/2939672.2939785}.
\newblock URL \url{https://doi.org/10.1145/2939672.2939785}.

\bibitem[Cohen et~al.(2024)Cohen, Khwaja, Wang, Masson, Ram{\'e}, Doubli, and
  Abou-Amal]{toto}
Ben Cohen, Emaad Khwaja, Kan Wang, Charles Masson, Elise Ram{\'e}, Youssef
  Doubli, and Othmane Abou-Amal.
\newblock Toto: Time series optimized transformer for observability.
\newblock \emph{arXiv preprint arXiv:2407.07874}, 2024.
\newblock URL \url{https://arxiv.org/abs/2407.07874}.

\bibitem[Cohen et~al.(2025)Cohen, Khwaja, Doubli, Lemaachi, Lettieri, Masson,
  Miccinilli, Ram{\'e}, Ren, Rostamizadeh, du~Terrail, Toon, Wang, Xie, Xu,
  Zhukova, Asker, Talwalkar, and Abou-Amal]{boom_data}
Ben Cohen, Emaad Khwaja, Youssef Doubli, Salahidine Lemaachi, Chris Lettieri,
  Charles Masson, Hugo Miccinilli, Elise Ram{\'e}, Qiqi Ren, Afshin
  Rostamizadeh, Jean du~Terrail, Anna-Monica Toon, Kan Wang, Stephan Xie,
  Zongzhe Xu, Viktoriya Zhukova, David Asker, Ameet Talwalkar, and Othmane
  Abou-Amal.
\newblock This time is different: An observability perspective on time series
  foundation models.
\newblock In \emph{Advances in Neural Information Processing Systems},
  volume~38, pp.\  50907--50951, 2025.
\newblock \doi{10.52202/085713-1698}.
\newblock URL
  \url{https://papers.neurips.cc/paper_files/paper/2025/hash/48ea942b60e6c52b50ca90e4a40d8ac2-Abstract-Conference.html}.

\bibitem[Comanici et~al.(2025)Comanici, Bieber, Schaekermann, Pasupat,
  Sachdeva, Dhillon, et~al.]{gemini25}
Gheorghe Comanici, Eric Bieber, Mike Schaekermann, Ice Pasupat, Noveen
  Sachdeva, Inderjit Dhillon, et~al.
\newblock {Gemini 2.5}: Pushing the frontier with advanced reasoning,
  multimodality, long context, and next generation agentic capabilities.
\newblock \emph{arXiv preprint arXiv:2507.06261}, 2025.
\newblock \doi{10.48550/arXiv.2507.06261}.

\bibitem[Das et~al.(2024)Das, Kong, Sen, and Zhou]{timesfm}
Abhimanyu Das, Weihao Kong, Rajat Sen, and Yichen Zhou.
\newblock A decoder-only foundation model for time-series forecasting.
\newblock In \emph{International Conference on Machine Learning}, 2024.
\newblock URL \url{https://arxiv.org/abs/2310.10688}.

\bibitem[Das et~al.(2026)Das, Goyal, Parmar, Song, Le, Miculicich, Yoon, Zhang,
  Palangi, and Pfister]{synapse}
Sarkar Snigdha~Sarathi Das, Palash Goyal, Mihir Parmar, Yiwen Song, Long Le,
  Lesly Miculicich, Jinsung Yoon, Rui Zhang, Hamid Palangi, and Tomas Pfister.
\newblock {Synapse}: Adaptive arbitration of complementary expertise in time
  series foundational models.
\newblock \emph{Transactions on Machine Learning Research}, 2026.
\newblock ISSN 2835-8856.
\newblock URL \url{https://openreview.net/forum?id=j3HqbsCwt1}.

\bibitem[{Datadog}(2026)]{totofnf}
{Datadog}.
\newblock {Toto-2.0-Family-and-Friends}.
\newblock Hugging Face model card and reproduction artifacts, 2026.
\newblock URL \url{https://huggingface.co/Datadog/Toto-2.0-Family-and-Friends}.

\bibitem[Ding et~al.(2026)Ding, Zhang, Dai, Wang, Zong, Liu, and
  Chu]{llatisa_data}
Yueyang Ding, HaoPeng Zhang, Rui Dai, Yi~Wang, Tianyu Zong, Kaikui Liu, and
  Xiangxiang Chu.
\newblock {LLaTiSA}: Towards difficulty-stratified time series reasoning from
  visual perception to semantics.
\newblock In \emph{Findings of the Association for Computational Linguistics:
  ACL 2026}, pp.\  32677--32717, 2026.
\newblock \doi{10.18653/v1/2026.findings-acl.1636}.
\newblock URL \url{https://aclanthology.org/2026.findings-acl.1636/}.

\bibitem[Gao et~al.(2024)Gao, Koker, Queen, Hartvigsen, Tsiligkaridis, and
  Zitnik]{units}
Shanghua Gao, Teddy Koker, Owen Queen, Thomas Hartvigsen, Theodoros
  Tsiligkaridis, and Marinka Zitnik.
\newblock {UniTS}: A unified multi-task time series model.
\newblock In \emph{Advances in Neural Information Processing Systems}, 2024.
\newblock URL \url{https://arxiv.org/abs/2403.00131}.

\bibitem[{Google Research}(2025)]{timesfm25}
{Google Research}.
\newblock {TimesFM 2.5}: {200M} {PyTorch} checkpoint.
\newblock Hugging Face model card, 2025.
\newblock URL \url{https://huggingface.co/google/timesfm-2.5-200m-pytorch}.

\bibitem[Goswami et~al.(2024)Goswami, Szafer, Choudhry, Cai, Li, and
  Dubrawski]{moment}
Mononito Goswami, Konrad Szafer, Arjun Choudhry, Yifu Cai, Shuo Li, and Artur
  Dubrawski.
\newblock {MOMENT}: A family of open time-series foundation models.
\newblock In \emph{International Conference on Machine Learning}, 2024.
\newblock URL \url{https://arxiv.org/abs/2402.03885}.

\bibitem[Graf et~al.(2026)Graf, Ortner, Wo{\'z}niak, and Pantazi]{flowstate}
Lars Graf, Thomas Ortner, Stanis{\l}aw Wo{\'z}niak, and Angeliki Pantazi.
\newblock {FlowState}: Sampling-rate-equivariant time-series forecasting.
\newblock In \emph{International Conference on Machine Learning}, 2026.
\newblock URL \url{https://arxiv.org/abs/2508.05287}.

\bibitem[Grattafiori et~al.(2024)Grattafiori, Dubey, Jauhri, Pandey, Kadian,
  Al-Dahle, et~al.]{llama3}
Aaron Grattafiori, Abhimanyu Dubey, Abhinav Jauhri, Abhinav Pandey, Abhishek
  Kadian, Ahmad Al-Dahle, et~al.
\newblock The {Llama 3} herd of models.
\newblock \emph{arXiv preprint arXiv:2407.21783}, 2024.
\newblock \doi{10.48550/arXiv.2407.21783}.

\bibitem[Gruver et~al.(2023)Gruver, Finzi, Qiu, and Wilson]{llmtime}
Nate Gruver, Marc Finzi, Shikai Qiu, and Andrew~Gordon Wilson.
\newblock Large language models are zero-shot time series forecasters.
\newblock In \emph{Advances in Neural Information Processing Systems}, 2023.
\newblock URL \url{https://arxiv.org/abs/2310.07820}.

\bibitem[Gu et~al.(2026)Gu, Zhao, Jing, and Ren]{tadiff_data}
Shuqi Gu, Yongxiang Zhao, Baoyu Jing, and Kan Ren.
\newblock What if tomorrow is the {World Cup} final? Counterfactual time series
  forecasting with textual conditions.
\newblock In \emph{International Conference on Machine Learning}, 2026.
\newblock URL \url{https://arxiv.org/abs/2605.14422}.

\bibitem[Guan et~al.(2026{\natexlab{a}})Guan, Meng, Li, Wang, Yang, Wen, Liu,
  Siniscalchi, Jin, and Pan]{timeomni}
Tong Guan, Zijie Meng, Dianqi Li, Shiyu Wang, Chao-Han~Huck Yang, Qingsong Wen,
  Zuozhu Liu, Sabato~Marco Siniscalchi, Ming Jin, and Shirui Pan.
\newblock {TimeOmni-1}: Incentivizing complex reasoning with time series in
  large language models.
\newblock In \emph{International Conference on Learning Representations},
  2026{\natexlab{a}}.
\newblock URL \url{https://arxiv.org/abs/2509.24803}.

\bibitem[Guan et~al.(2026{\natexlab{b}})Guan, Pan, Barthelemy, Li, Cai, Alippi,
  Jin, and Pan]{timeomnivl}
Tong Guan, Sheng Pan, Johan Barthelemy, Zhao Li, Yujun Cai, Cesare Alippi, Ming
  Jin, and Shirui Pan.
\newblock {TimeOmni-VL}: Unified models for time series understanding and
  generation.
\newblock In \emph{International Conference on Machine Learning},
  2026{\natexlab{b}}.
\newblock URL \url{https://arxiv.org/abs/2602.17149}.

\bibitem[Gwiazda et~al.(2026)Gwiazda, Cai, Goswami, Choudhry, and
  Dubrawski]{tseagent}
Malgorzata Gwiazda, Yifu Cai, Mononito Goswami, Arjun Choudhry, and Artur
  Dubrawski.
\newblock {TimeSeriesExamAgent}: Creating time series reasoning benchmarks at
  scale.
\newblock In \emph{International Conference on Learning Representations}, 2026.
\newblock URL
  \url{https://proceedings.iclr.cc/paper_files/paper/2026/hash/8b6dd6597c2edeb9ee62468ccea6f512-Abstract-Conference.html}.

\bibitem[Hu et~al.(2022)Hu, Shen, Wallis, Allen-Zhu, Li, Wang, Wang, and
  Chen]{lora}
Edward~J. Hu, Yelong Shen, Phillip Wallis, Zeyuan Allen-Zhu, Yuanzhi Li, Shean
  Wang, Lu~Wang, and Weizhu Chen.
\newblock {LoRA}: Low-rank adaptation of large language models.
\newblock In \emph{International Conference on Learning Representations}, 2022.
\newblock URL \url{https://arxiv.org/abs/2106.09685}.

\bibitem[{Iowa Environmental Mesonet}(n.d.)]{nws_archive}
{Iowa Environmental Mesonet}.
\newblock {NWS} text product finder, n.d.
\newblock URL \url{https://mesonet.agron.iastate.edu/wx/afos/}.
\newblock Accessed September 21, 2026.

\bibitem[Jain \& Sen(2026)Jain and Sen]{timesfm3}
Ayush Jain and Rajat Sen.
\newblock {TimesFM-3}: A zero-shot foundation model for multivariate
  forecasting.
\newblock Google Research, 2026.
\newblock URL
  \url{https://research.google/blog/timesfm-3-a-zero-shot-foundation-model-for-multivariate-forecasting/}.

\bibitem[Jin et~al.(2024)Jin, Wang, Ma, Chu, Zhang, Shi, Chen, Liang, Li, Pan,
  and Wen]{timellm}
Ming Jin, Shiyu Wang, Lintao Ma, Zhixuan Chu, James~Y. Zhang, Xiaoming Shi,
  Pin-Yu Chen, Yuxuan Liang, Yuan-Fang Li, Shirui Pan, and Qingsong Wen.
\newblock {Time-LLM}: Time series forecasting by reprogramming large language
  models.
\newblock In \emph{International Conference on Learning Representations}, 2024.
\newblock URL \url{https://arxiv.org/abs/2310.01728}.

\bibitem[Kayaalp et~al.(2026)Kayaalp, Turkmen, Shchur, Mercado, Ansari,
  Bohlke-Schneider, and Wang]{chroma}
Mert Kayaalp, Caner Turkmen, Oleksandr Shchur, Pedro Mercado, Abdul~Fatir
  Ansari, Michael Bohlke-Schneider, and Bernie Wang.
\newblock Test-time efficient pretrained model portfolios for time series
  forecasting.
\newblock In \emph{International Conference on Learning Representations}, pp.\
  16739--16772, 2026.
\newblock URL
  \url{https://proceedings.iclr.cc/paper_files/paper/2026/file/1c1b099a19621e4bd2753ac572e1dbd5-Paper-Conference.pdf}.

\bibitem[Khwaja et~al.(2026)Khwaja, Lettieri, Woo, Belouadah, Cenac, Jarry,
  Paquin, Zhao, Zhukova, Abou-Amal, Liu, Talwalkar, and Asker]{toto2}
Emaad Khwaja, Chris Lettieri, Gerald Woo, Eden Belouadah, Marc Cenac, Guillaume
  Jarry, Enguerrand Paquin, Xunyi Zhao, Viktoriya Zhukova, Othmane Abou-Amal,
  Chenghao Liu, Ameet Talwalkar, and David Asker.
\newblock {Toto 2.0}: Time series forecasting enters the scaling era.
\newblock \emph{arXiv preprint arXiv:2605.20119}, 2026.
\newblock URL \url{https://arxiv.org/abs/2605.20119}.

\bibitem[Kong et~al.(2025)Kong, Yang, Hwang, Du, Zohren, Wang, Jin, and
  Wen]{timemqa_data}
Yaxuan Kong, Yiyuan Yang, Yoontae Hwang, Wenjie Du, Stefan Zohren, Zhangyang
  Wang, Ming Jin, and Qingsong Wen.
\newblock {Time-MQA}: Time series multi-task question answering with context
  enhancement.
\newblock In \emph{Proceedings of the 63rd Annual Meeting of the Association
  for Computational Linguistics}, pp.\  29736--29753, 2025.
\newblock \doi{10.18653/v1/2025.acl-long.1437}.
\newblock URL \url{https://aclanthology.org/2025.acl-long.1437/}.

\bibitem[Langer et~al.(2026)Langer, Kaar, Rosenblattl, Xu, Chow, Maritsch,
  Jakob, Wang, Liu, Verma, Han, Kim, Chubb, Ceresnak, Zahedivash, Sandhu,
  Rodriguez, McDuff, Fleisch, Aalami, Barata, and Schmiedmayer]{opentslm}
Patrick Langer, Thomas Kaar, Max Rosenblattl, Maxwell~A. Xu, Winnie Chow,
  Martin Maritsch, Robert Jakob, Ning Wang, Juncheng Liu, Aradhana Verma, Brian
  Han, Daniel~Seung Kim, Henry Chubb, Scott Ceresnak, Aydin Zahedivash,
  Alexander Tarlochan~Singh Sandhu, Fatima Rodriguez, Daniel McDuff, Elgar
  Fleisch, Oliver Aalami, Filipe Barata, and Paul Schmiedmayer.
\newblock {OpenTSLM}: Time-series language models for reasoning over
  multivariate medical text- and time-series data.
\newblock In \emph{Proceedings of the 43rd International Conference on Machine
  Learning}, 2026.
\newblock URL \url{https://icml.cc/virtual/2026/poster/65261}.

\bibitem[Li et~al.(2026)Li, Jin, Bei, Zou, Kumar, Ning, Zhao, Ai, Jing, Tong,
  and He]{timeclaw}
Zihao Li, Kaifeng Jin, Yuanchen Bei, Jiaru Zou, Avaneesh Kumar, Xuying Ning,
  Yanjun Zhao, Mengting Ai, Baoyu Jing, Hanghang Tong, and Jingrui He.
\newblock Harnessing generalist agents for contextualized time series.
\newblock \emph{arXiv preprint arXiv:2606.05404}, 2026.
\newblock URL \url{https://arxiv.org/abs/2606.05404}.

\bibitem[Liu et~al.(2025)Liu, Aksu, Liu, Liu, Yan, Pham, Savarese, Sahoo,
  Xiong, and Li]{moirai2}
Chenghao Liu, Taha Aksu, Juncheng Liu, Xu~Liu, Hanshu Yan, Quang Pham, Silvio
  Savarese, Doyen Sahoo, Caiming Xiong, and Junnan Li.
\newblock {Moirai 2.0}: When less is more for time series forecasting.
\newblock \emph{arXiv preprint arXiv:2511.11698}, 2025.
\newblock URL \url{https://arxiv.org/abs/2511.11698}.

\bibitem[Liu et~al.(2026{\natexlab{a}})Liu, Fons, Vapsi, Ghassemi, Vyetrenko,
  Borrajo, Potluru, and Veloso]{tsagent}
Penghang Liu, Elizabeth Fons, Annita Vapsi, Mohsen Ghassemi, Svitlana
  Vyetrenko, Daniel Borrajo, Vamsi~K. Potluru, and Manuela Veloso.
\newblock {TS-Agent}: Understanding and reasoning over raw time series via
  iterative insight gathering.
\newblock \emph{arXiv preprint arXiv:2510.07432}, 2026{\natexlab{a}}.
\newblock URL \url{https://arxiv.org/abs/2510.07432}.

\bibitem[Liu et~al.(2024)Liu, Zhang, Li, Huang, Wang, and Long]{timer}
Yong Liu, Haoran Zhang, Chenyu Li, Xiangdong Huang, Jianmin Wang, and Mingsheng
  Long.
\newblock Timer: Generative pre-trained transformers are large time series
  models.
\newblock In \emph{International Conference on Machine Learning}, 2024.
\newblock URL \url{https://arxiv.org/abs/2402.02368}.

\bibitem[Liu et~al.(2026{\natexlab{b}})Liu, Su, Wang, Zhang, Liu, Wang, Ye,
  Xiang, Wang, and Long]{timers1}
Yong Liu, Xingjian Su, Shiyu Wang, Haoran Zhang, Haixuan Liu, Yuxuan Wang, Zhou
  Ye, Yang Xiang, Jianmin Wang, and Mingsheng Long.
\newblock {Timer-S1}: A billion-scale time series foundation model with serial
  scaling.
\newblock \emph{arXiv preprint arXiv:2603.04791}, 2026{\natexlab{b}}.
\newblock URL \url{https://arxiv.org/abs/2603.04791}.

\bibitem[Montero-Manso et~al.(2020)Montero-Manso, Athanasopoulos, Hyndman, and
  Talagala]{fforma}
Pablo Montero-Manso, George Athanasopoulos, Rob~J. Hyndman, and Thiyanga~S.
  Talagala.
\newblock {FFORMA}: Feature-based forecast model averaging.
\newblock \emph{International Journal of Forecasting}, 36\penalty0
  (1):\penalty0 86--92, 2020.
\newblock \doi{10.1016/j.ijforecast.2019.02.011}.
\newblock URL \url{https://robjhyndman.com/publications/fforma/}.

\bibitem[Nguyen et~al.(2026)Nguyen, Nguyen, Nguyen, Do, and Le]{lafp}
Huu~Hiep Nguyen, Dung Nguyen, Minh~Hoang Nguyen, Dai Do, and Hung Le.
\newblock {LLM} as forecasting planner: Training-free text conditioning for
  time-series foundation models.
\newblock \emph{arXiv preprint arXiv:2607.24892}, 2026.
\newblock URL \url{https://arxiv.org/abs/2607.24892}.

\bibitem[{OpenAI}(2024)]{gpt4o}
{OpenAI}.
\newblock {GPT-4o System Card}.
\newblock \emph{arXiv preprint arXiv:2410.21276}, 2024.
\newblock \doi{10.48550/arXiv.2410.21276}.

\bibitem[{OpenAI}(2025{\natexlab{a}})]{gpt52}
{OpenAI}.
\newblock Update to {GPT-5} system card: {GPT-5.2}.
\newblock System card, 2025{\natexlab{a}}.
\newblock URL \url{https://openai.com/index/gpt-5-system-card-update-gpt-5-2/}.

\bibitem[{OpenAI}(2025{\natexlab{b}})]{gptoss}
{OpenAI}.
\newblock {gpt-oss-120b \& gpt-oss-20b Model Card}.
\newblock \emph{arXiv preprint arXiv:2508.10925}, 2025{\natexlab{b}}.
\newblock \doi{10.48550/arXiv.2508.10925}.

\bibitem[Ostapenko et~al.(2024)Ostapenko, Su, Ponti, Charlin, Le~Roux, Caccia,
  and Sordoni]{ostapenko2024modular}
Oleksiy Ostapenko, Zhan Su, Edoardo Ponti, Laurent Charlin, Nicolas Le~Roux,
  Lucas Caccia, and Alessandro Sordoni.
\newblock Towards modular {LLM}s by building and reusing a library of {LoRA}s.
\newblock In \emph{Proceedings of the 41st International Conference on Machine
  Learning}, volume 235 of \emph{Proceedings of Machine Learning Research},
  pp.\  38885--38904. PMLR, 2024.
\newblock URL \url{https://proceedings.mlr.press/v235/ostapenko24a.html}.

\bibitem[Parker et~al.(2026)Parker, Chan, Zhang, and Ghobadi]{tsllm}
Felix Parker, Nimeesha Chan, Chi Zhang, and Kimia Ghobadi.
\newblock {TsLLM}: Augmenting {LLMs} for general time series understanding and
  prediction.
\newblock In \emph{International Conference on Machine Learning}, 2026.
\newblock URL \url{https://icml.cc/virtual/2026/poster/61098}.
\newblock Protocol comparisons refer to arXiv:2510.01111v2,
  \url{https://arxiv.org/abs/2510.01111v2}.

\bibitem[Pfeiffer et~al.(2021)Pfeiffer, Kamath, R{\"u}ckl{\'e}, Cho, and
  Gurevych]{pfeiffer-etal-2021-adapterfusion}
Jonas Pfeiffer, Aishwarya Kamath, Andreas R{\"u}ckl{\'e}, Kyunghyun Cho, and
  Iryna Gurevych.
\newblock {AdapterFusion}: Non-destructive task composition for transfer
  learning.
\newblock In \emph{Proceedings of the 16th Conference of the European Chapter
  of the Association for Computational Linguistics: Main Volume}, pp.\
  487--503. Association for Computational Linguistics, 2021.
\newblock \doi{10.18653/v1/2021.eacl-main.39}.
\newblock URL \url{https://aclanthology.org/2021.eacl-main.39/}.

\bibitem[Pierrot \& Pinson(2024)Pierrot and Pinson]{pierrot_pinson_data}
Amandine Pierrot and Pierre Pinson.
\newblock On tracking varying bounds when forecasting bounded time series.
\newblock \emph{Technometrics}, 66\penalty0 (4):\penalty0 651--661, 2024.
\newblock \doi{10.1080/00401706.2024.2350421}.
\newblock URL \url{https://doi.org/10.1080/00401706.2024.2350421}.

\bibitem[Podest et~al.(2026)Podest, Pichler, B{\"u}rger, Z{\'o}lyomi,
  Voggenberger, Berghammer, Klotz, B{\"o}ck, Klambauer, and Hochreiter]{tirex2}
Patrick Podest, Marco Pichler, Elias B{\"u}rger, Levente Z{\'o}lyomi, Bernhard
  Voggenberger, Wilhelm Berghammer, Daniel Klotz, Sebastian B{\"o}ck,
  G{\"u}nter Klambauer, and Sepp Hochreiter.
\newblock {TiRex-2}: Generalizing {TiRex} to multivariate data and streaming.
\newblock \emph{arXiv preprint arXiv:2607.01204}, 2026.
\newblock URL \url{https://arxiv.org/abs/2607.01204}.

\bibitem[{Qwen Team}(2025)]{qwen3next}
{Qwen Team}.
\newblock {Qwen3-Next-80B-A3B-Instruct}.
\newblock Hugging Face model card, 2025.
\newblock URL \url{https://huggingface.co/Qwen/Qwen3-Next-80B-A3B-Instruct}.

\bibitem[{Qwen Team}(2026)]{qwen36_27b}
{Qwen Team}.
\newblock {Qwen3.6-27B}: Flagship-level coding in a {27B} dense model, April
  2026.
\newblock URL \url{https://qwen.ai/blog?id=qwen3.6-27b}.

\bibitem[{Racine.ai}(2026)]{racinecast}
{Racine.ai}.
\newblock {RacineCast-1}.
\newblock Hugging Face model card and reproduction artifacts, 2026.
\newblock URL \url{https://huggingface.co/racineai/RacineCast-1}.

\bibitem[Rasul et~al.(2023)Rasul, Ashok, Williams, Ghonia, Bhagwatkar,
  Khorasani, Bayazi, Adamopoulos, Riachi, Hassen, Bilo{\v{s}}, Garg, Schneider,
  Chapados, Drouin, Zantedeschi, Nevmyvaka, and Rish]{lagllama}
Kashif Rasul, Arjun Ashok, Andrew~Robert Williams, Hena Ghonia, Rishika
  Bhagwatkar, Arian Khorasani, Mohammad Javad~Darvishi Bayazi, George
  Adamopoulos, Roland Riachi, Nadhir Hassen, Marin Bilo{\v{s}}, Sahil Garg,
  Anderson Schneider, Nicolas Chapados, Alexandre Drouin, Valentina
  Zantedeschi, Yuriy Nevmyvaka, and Irina Rish.
\newblock {Lag-Llama}: Towards foundation models for probabilistic time series
  forecasting.
\newblock \emph{arXiv preprint arXiv:2310.08278}, 2023.
\newblock URL \url{https://arxiv.org/abs/2310.08278}.

\bibitem[Rechtor{\'i}k et~al.(2026)Rechtor{\'i}k, Du{\v{s}}ek, and
  Kasner]{codingagents}
Filip Rechtor{\'i}k, Ond{\v{r}}ej Du{\v{s}}ek, and Zden{\v{e}}k Kasner.
\newblock Can {LLM} coding agents reason about time series?
\newblock \emph{arXiv preprint arXiv:2606.16545}, 2026.
\newblock URL \url{https://arxiv.org/abs/2606.16545}.

\bibitem[{Salesforce}(2026)]{gifteval_leaderboard}
{Salesforce}.
\newblock {GIFT-Eval}: Official leaderboard implementation.
\newblock Hugging Face Space, 2026.
\newblock URL
  \url{https://huggingface.co/spaces/Salesforce/GIFT-Eval/blob/main/app.py}.
\newblock Accessed September 23, 2026.

\bibitem[{Salesforce AI Research}(2026)]{gifteval_results}
{Salesforce AI Research}.
\newblock {GIFT-Eval}: Official results snapshot.
\newblock GitHub repository, 2026.
\newblock URL
  \url{https://github.com/SalesforceAIResearch/gift-eval/tree/main/results}.
\newblock Accessed September 23, 2026.

\bibitem[Sun et~al.(2026)Sun, Tian, Huang, Feng, and Zhang]{rmisc_data}
Qian Sun, Yong-Ming Tian, Jia-Wei Huang, Cheng Feng, and Shao-Qun Zhang.
\newblock {RMISC}: A large-scale real-world multivariate corpus for time series
  foundation models.
\newblock \emph{arXiv preprint arXiv:2607.06504}, 2026.
\newblock URL \url{https://arxiv.org/abs/2607.06504}.

\bibitem[Tang et~al.(2026)Tang, Williams, Ashok, Zheng, Sun, Drouin, Laradji,
  Marcotte, and Zantedeschi]{drcik}
Yihong Tang, Andrew~Robert Williams, Arjun Ashok, Vincent~Zhihao Zheng, Lijun
  Sun, Alexandre Drouin, Issam~H. Laradji, {\'E}tienne Marcotte, and Valentina
  Zantedeschi.
\newblock {Dr-CiK}: A testbed for foresight-driven agents.
\newblock \emph{arXiv preprint arXiv:2605.27904}, 2026.
\newblock URL \url{https://arxiv.org/abs/2605.27904}.

\bibitem[Wang et~al.(2025{\natexlab{a}})Wang, Qi, Wang, Sun, Zhuang, Wu, Zhang,
  and Liao]{chattime_data}
Chengsen Wang, Qi~Qi, Jingyu Wang, Haifeng Sun, Zirui Zhuang, Jinming Wu, Lei
  Zhang, and Jianxin Liao.
\newblock {ChatTime}: A unified multimodal time series foundation model
  bridging numerical and textual data.
\newblock In \emph{Proceedings of the AAAI Conference on Artificial
  Intelligence}, volume~39, pp.\  12694--12702, 2025{\natexlab{a}}.
\newblock \doi{10.1609/aaai.v39i12.33384}.
\newblock URL \url{https://doi.org/10.1609/aaai.v39i12.33384}.

\bibitem[Wang et~al.(2025{\natexlab{b}})Wang, Lei, Song, Hao, Chen, Zhang, Jia,
  Li, and Wei]{itformer}
Yilin Wang, Peixuan Lei, Jie Song, Yuzhe Hao, Tao Chen, Yuxuan Zhang, Lei Jia,
  Yuanxiang Li, and Zhongyu Wei.
\newblock {ITFormer}: Bridging time series and natural language for multi-modal
  {QA} with large-scale multitask dataset.
\newblock In \emph{International Conference on Machine Learning}, volume 267 of
  \emph{Proceedings of Machine Learning Research}, pp.\  63324--63344. PMLR,
  2025{\natexlab{b}}.
\newblock URL \url{https://proceedings.mlr.press/v267/wang25av.html}.

\bibitem[Wen et~al.(2026)Wen, Gifford, Reddy, Nguyen, Kalagnanam, and
  Julius]{patchtstfm}
Yunshi Wen, Wesley~M. Gifford, Chandra Reddy, Lam~M. Nguyen, Jayant Kalagnanam,
  and Anak~Agung Julius.
\newblock Revisiting the generic transformer: Deconstructing a strong baseline
  for time series foundation models.
\newblock \emph{arXiv preprint arXiv:2602.06909}, 2026.
\newblock URL \url{https://arxiv.org/abs/2602.06909}.

\bibitem[Williams et~al.(2025)Williams, Ashok, Marcotte, Zantedeschi,
  Subramanian, Riachi, Requeima, Lacoste, Rish, Chapados, and
  Drouin]{contextiskey}
Andrew~Robert Williams, Arjun Ashok, {\'E}tienne Marcotte, Valentina
  Zantedeschi, Jithendaraa Subramanian, Roland Riachi, James Requeima,
  Alexandre Lacoste, Irina Rish, Nicolas Chapados, and Alexandre Drouin.
\newblock {Context is Key}: A benchmark for forecasting with essential textual
  information.
\newblock In \emph{International Conference on Machine Learning}, 2025.
\newblock URL \url{https://arxiv.org/abs/2410.18959}.

\bibitem[Woo et~al.(2024)Woo, Liu, Kumar, Xiong, Savarese, and Sahoo]{moirai}
Gerald Woo, Chenghao Liu, Akshat Kumar, Caiming Xiong, Silvio Savarese, and
  Doyen Sahoo.
\newblock Unified training of universal time series forecasting transformers.
\newblock In \emph{International Conference on Machine Learning}, 2024.
\newblock URL \url{https://arxiv.org/abs/2402.02592}.

\bibitem[Wu et~al.(2024)Wu, Huang, and Wei]{wu2024mole}
Xun Wu, Shaohan Huang, and Furu Wei.
\newblock Mixture of {LoRA} experts.
\newblock In \emph{International Conference on Learning Representations}, 2024.
\newblock URL
  \url{https://proceedings.iclr.cc/paper_files/paper/2024/hash/ce806d8b4bf38cd92d483e5a0490d983-Abstract-Conference.html}.

\bibitem[Xie et~al.(2026)Xie, Cohen, Goswami, Shen, Khwaja, Liu, Asker,
  Abou-Amal, and Talwalkar]{arfbench}
Stephan Xie, Ben Cohen, Mononito Goswami, Junhong Shen, Emaad Khwaja, Chenghao
  Liu, David Asker, Othmane Abou-Amal, and Ameet Talwalkar.
\newblock {ARFBench}: Benchmarking time series question answering ability for
  software incident response.
\newblock \emph{arXiv preprint arXiv:2604.21199}, 2026.
\newblock URL \url{https://arxiv.org/abs/2604.21199}.

\bibitem[Xie et~al.(2025)Xie, Li, He, Xu, Wen, Zhang, Chen, Shi, and
  Pei]{chatts}
Zhe Xie, Zeyan Li, Xiao He, Longlong Xu, Xidao Wen, Tieying Zhang, Jianjun
  Chen, Rui Shi, and Dan Pei.
\newblock {ChatTS}: Aligning time series with {LLMs} via synthetic data for
  enhanced understanding and reasoning.
\newblock \emph{Proceedings of the VLDB Endowment}, 18\penalty0 (8):\penalty0
  2385--2398, 2025.
\newblock \doi{10.14778/3742728.3742735}.
\newblock URL \url{https://www.vldb.org/pvldb/vol18/p2385-xie.pdf}.

\bibitem[Xue \& Salim(2024)Xue and Salim]{promptcast}
Hao Xue and Flora~D. Salim.
\newblock {PromptCast}: A new prompt-based learning paradigm for time series
  forecasting.
\newblock \emph{IEEE Transactions on Knowledge and Data Engineering},
  36\penalty0 (11):\penalty0 6851--6864, 2024.
\newblock \doi{10.1109/TKDE.2023.3342137}.
\newblock URL \url{https://doi.org/10.1109/TKDE.2023.3342137}.

\bibitem[Ye et~al.(2026)Ye, Yang, Cao, Zhang, Tang, Cai, and
  Liu]{domainreasoner}
Wen Ye, Wei Yang, Defu Cao, Yizhou Zhang, Lumingyuan Tang, Jie Cai, and Yan
  Liu.
\newblock {TS-Reasoner}: Domain-oriented time series inference agents for
  reasoning and automated analysis.
\newblock \emph{Transactions on Machine Learning Research}, 2026.
\newblock URL \url{https://openreview.net/forum?id=yhy7Vigjcf}.

\bibitem[Yu et~al.(2026)Yu, Zhao, and Zhou]{tsreasoner}
Fangxu Yu, Hongyu Zhao, and Tianyi Zhou.
\newblock {TS-Reasoner}: Aligning time series foundation models with {LLM}
  reasoning.
\newblock \emph{Transactions on Machine Learning Research}, 2026.
\newblock ISSN 2835-8856.
\newblock URL \url{https://openreview.net/forum?id=d6TD0f2xXq}.

\bibitem[Zhang et~al.(2026{\natexlab{a}})Zhang, Han, Ansari, Auer, Wang,
  Goswami, Zhang, Wang, Bohlke-Schneider, and Wang]{codecast}
Xiyuan Zhang, Boran Han, Abdul~Fatir Ansari, Andreas Auer, Rui Wang, Mononito
  Goswami, Alexander Zhang, Hao Wang, Michael Bohlke-Schneider, and Yuyang
  Wang.
\newblock {CodeCast}: Context-conditional code generation for multimodal time
  series forecasting.
\newblock In \emph{ICML 2026 Workshop on Deep Learning for Code: Towards
  Human-Centered Coding Agents}, 2026{\natexlab{a}}.
\newblock URL \url{https://icml.cc/virtual/2026/82732}.

\bibitem[Zhang et~al.(2026{\natexlab{b}})Zhang, Han, Fang, Ansari, Zhang,
  Maddix, Hu, Wilson, Mahoney, Wang, Liu, Bohlke-Schneider, Rangwala, Karypis,
  and Wang]{multimodalityforecasting}
Xiyuan Zhang, Boran Han, Haoyang Fang, Abdul~Fatir Ansari, Shuai Zhang,
  Danielle~C. Maddix, Cuixiong Hu, Andrew~Gordon Wilson, Michael~W. Mahoney,
  Hao Wang, Yan Liu, Michael Bohlke-Schneider, Huzefa Rangwala, George Karypis,
  and Bernie Wang.
\newblock When does multimodality lead to better time series forecasting?
\newblock \emph{Transactions on Machine Learning Research}, 2026{\natexlab{b}}.
\newblock ISSN 2835-8856.
\newblock URL \url{https://openreview.net/forum?id=RggcWYWR3N}.
\newblock Preprint version: arXiv:2506.21611v2.

\bibitem[Zheng et~al.(2026)Zheng, Marcotte, Ashok, Williams, Sun, Drouin, and
  Zantedeschi]{caf_data}
Vincent~Zhihao Zheng, {\'E}tienne Marcotte, Arjun Ashok, Andrew~Robert
  Williams, Lijun Sun, Alexandre Drouin, and Valentina Zantedeschi.
\newblock Overcoming the modality gap in context-aided forecasting.
\newblock In \emph{International Conference on Machine Learning}, 2026.
\newblock URL \url{https://arxiv.org/abs/2603.12451}.

\bibitem[Zhou et~al.(2026)Zhou, Li, Li, Zhang, Meng, Li, Chen, Lou, and
  Ng]{veritime}
Jiahui Zhou, Dan Li, Boxin Li, Xiao Zhang, Erli Meng, Lin Li, Zhuomin Chen,
  Jian Lou, and See-Kiong Ng.
\newblock Time series reasoning via process-verifiable thinking data synthesis
  and scheduling for tailored {LLM} reasoning.
\newblock In \emph{International Conference on Machine Learning}, 2026.
\newblock URL \url{https://arxiv.org/abs/2602.07830v2}.

\bibitem[Zhou et~al.(2023)Zhou, Niu, Wang, Sun, and Jin]{gpt4ts}
Tian Zhou, Peisong Niu, Xue Wang, Liang Sun, and Rong Jin.
\newblock One fits all: Power general time series analysis by pretrained {LM}.
\newblock In \emph{Advances in Neural Information Processing Systems}, 2023.
\newblock URL \url{https://arxiv.org/abs/2302.11939}.

\end{thebibliography}
\end{document}